\documentclass[final]{cvpr}

\usepackage{times}
\usepackage{epsfig}
\usepackage{graphicx}
\usepackage{amsmath}
\usepackage{amssymb}
\usepackage[utf8]{inputenc} % allow utf-8 input
\usepackage[T1]{fontenc}    % use 8-bit T1 fonts
\usepackage{booktabs}       % professional-quality tables
\usepackage{amsfonts}       % blackboard math symbols
\usepackage{nicefrac}       % compact symbols for 1/2, etc.
\usepackage{microtype}      % microtypography
\usepackage{caption}
\usepackage{subfigure}
\usepackage{amsthm}
\usepackage{url}
\usepackage{bm}
\usepackage{wrapfig}
\usepackage{epstopdf}

\usepackage[pagebackref=true,breaklinks=true,colorlinks,bookmarks=false]{hyperref}
\newtheorem{proposition}{Proposition}
\newtheorem{definition}{Definition}

\newcommand{\nnMass}{m}
\newcommand{\lj}{\bm{J_{i,i-1}}}

\newcommand{\rgKbar}{\bar{k}_{\mathcal{R}|\mathcal{G}}}

\def\cvprPaperID{7774} % *** Enter the CVPR Paper ID here
\def\confYear{CVPR 2021}

\begin{document}

%%%%%%%%% TITLE
\title{How does topology influence gradient propagation and\\model performance of deep networks with DenseNet-type skip connections?}
\author{Kartikeya Bhardwaj$^{1,*}$, Guihong Li$^{2,}$\thanks{Equal Contribution}\ , and Radu Marculescu$^2$\\
$^1$Arm Inc., San Jose, CA, USA 95134\\
$^2$The University of Texas at Austin, Austin, TX, USA 78712\\
{\tt\small kartikeya.bhardwaj@arm.com, \{lgh,radum\}@utexas.edu}
}

\maketitle
\thispagestyle{empty}
\pagestyle{empty}

%%%%%%%%% ABSTRACT
\begin{abstract}\vspace{-3mm}
DenseNets introduce concatenation-type skip connections that achieve state-of-the-art accuracy in several computer vision tasks. 
In this paper, we reveal that the topology of the concatenation-type skip connections is closely related to the gradient propagation which, in turn, enables a predictable behavior of DNNs' test performance.
To this end, we introduce a new metric called \textit{NN-Mass} to quantify how effectively information flows through DNNs. 
Moreover, we empirically show that NN-Mass also works for other types of skip connections, e.g., for ResNets, Wide-ResNets (WRNs), and MobileNets, which contain addition-type skip connections (i.e., residuals or inverted residuals). As such, for both DenseNet-like CNNs and ResNets/WRNs/MobileNets, our theoretically grounded NN-Mass can identify models with similar accuracy, despite having significantly different size/compute requirements. Detailed experiments on both synthetic and real datasets (\textit{e.g.}, MNIST, CIFAR-10, CIFAR-100, ImageNet) provide extensive evidence for our insights. Finally, the closed-form equation of our NN-Mass enables us to \textit{design} significantly compressed DenseNets (for CIFAR-10) and MobileNets (for ImageNet) directly at initialization without time-consuming training and/or searching.\footnote{Code
at \url{https://github.com/SLDGroup/NN_Mass}.}
\end{abstract}

%%%%%%%%% BODY TEXT
\vspace{-3mm}
\section{Introduction}\vspace{-2mm}
DenseNets~\cite{densenet} and their variants have been widely adopted by the deep learning community to achieve excellent performance in many computer vision tasks such as image classification, object detection, image segmentation, super resolution, among many others~\cite{dense_od, dense_seg, zhang2018residual}. One of the main contributions of DenseNets is the introduction of \textit{concatenation-type skip connections} where the output channels from all previous layers are concatenated at the input of the current convolutional layer. The concatenation-type skip connections\footnote{Also referred to as DenseNet-type skip connections.} have been particularly valuable to the deep learning literature. For instance, in addition to significant accuracy and efficiency gains in computer vision applications, many state-of-the-art Neural Architecture Search (NAS) techniques have exploited the concatenation-type skip connections into their search space to obtain high-performance models~\cite{enas, dense_nas1, real2019regularized, darts, nasnet}. However, to the best of our knowledge, the properties of concatenation-type skip connections such as their gradient propagation and the resulting effect on model performance has \textit{not} been explored. 

Recently, an important method called Dynamical Isometry emerged in order to quantify gradient flow through DNNs~\cite{saxe2013exact, tarnowski2018dynamical, pennington2017resurrecting, sigprop}. When DNNs achieve ``dynamical isometry'', the signal flows through such networks without significant amplification or attenuation. This, in turn, helps the learning process and, hence, quantifies the gradient properties of DNNs. The role of Dynamical Isometry in trainability has been demonstrated for networks such as ResNets~\cite{resnet} which have \textit{addition-type skip connections}.

Due to concatenation of channels, DenseNet-type skip connections enforce strong \textit{structural/topological} constraints on the gradient propagation (\textit{i.e.}, the gradients can only follow specific paths during training). In general, the topology (or structure) of graphs/networks directly influences the process taking place over them~\cite{networksci}. For instance, how closely the users of a social network are connected to each other completely determines how fast the information propagates through the network~\cite{leskovec2007patterns, kempe2003maximizing}. Consequently, the structural constraints imposed by concatenation of channels must also affect the learning dynamics of DNNs.
Motivated by this observation, we study the relationship between topology, gradient flow, and model performance of such deep networks. Note that, to study these topological properties, we do not use the original DenseNets which contain all-to-all connections~\cite{densenet}, but rather a generalized version where we can vary the density of skip connections (more details in Section~\ref{app}).

To this end, we first define our setup of DNNs with concatenation-type skip connections. Then, we propose a new metric called \textit{NN-Mass} to quantify the topological properties of DNNs considered within this setup. Next, we show the relationship between NN-Mass (a topological property) and Layerwise Dynamical Isometry (LDI)~\cite{sigprop}, a property that indicates the faithful gradient propagation through the network~\cite{saxe2013exact}. Specifically, we show that irrespective of number of parameters/FLOPS/layers, models with similar NN-Mass and width should have similar LDI, and thus a similar gradient flow that results in comparable accuracy. 

To support these theoretical insights, we conduct extensive experiments to show that models with the same width and NN-Mass indeed achieve a similar accuracy irrespective of their depth, number of parameters (\#Params), and FLOPS. 
Moreover, we empirically show that NN-Mass also works for other types of skip connections, e.g., for ResNets, Wide-ResNets (WRNs), and MobileNets which contain addition-type skip connections (ATSC), \textit{i.e.}, residuals or inverted residuals. 
Finally, we show how the closed-form expression for NN-Mass can be used to \textit{directly} design compressed DNNs, that is, without \textit{any} time-consuming training and (manual or automatic) searching for compressed models.  

Overall, we make the following \textbf{key contributions}: (\textit{i})~We reveal how topological constraints imposed by DenseNet-type skip connections influence gradient propagation and resulting accuracy; (\textit{ii})~For this setup, we propose a new topological metric called NN-Mass that is theoretically linked to Layerwise Dynamical Isometry and quantifies how efficiently information propagates in neural networks; (\textit{iii})~Our experiments encompass multilayer perceptron as well as CNNs with DenseNet-type skip connections on several datasets (MNIST, CIFAR-10, CIFAR-100, Imagenet). Our results demonstrate that NN-Mass is an excellent indicator of accuracy and support our theory. We further empirically show that NN-Mass also works for ATSC-based networks (ResNets, WRNs, and MobileNets); (\textit{iv})~Finally, NN-Mass allows us to directly design models with up to $3\times$ compression rate (for DenseNets on CIFAR-10), and up to $34\%$-$40\%$ compression rate (for MobileNet-v2 on ImageNet) in \#Params/FLOPS while losing minimal accuracy.

The rest of the paper is organized as follows: Section~\ref{rel} discusses the related work and some preliminaries. Then, Section~\ref{app} describes our proposed metric and its theoretical analysis. Section~\ref{exp} presents detailed experimental results. Finally, Section~\ref{conc} summarizes our work and contributions. 

\section{Background and Related Work}\label{rel}
Several prior works aim to study the impact of initialization on model convergence and gradients~\cite{lecun2012efficient, glorot2010understanding, saxe2013exact, poole2016exponential, tarnowski2018dynamical, pennington2017resurrecting}. To this end, Dynamical Isometry has emerged as an important metric for quantifying gradient properties. Moreover, recent model compression literature attempts to connect pruning at initialization to gradient properties~\cite{sigprop}. However, none of these studies address the impact of the \textit{topology} of concatenation-type skip connections on gradient propagation. Instead, since our objective is to specifically study the topological properties, we rely on graph theory/network concepts. Hence, our work is orthogonal to prior art that explores the impact of initialization on gradients as those works do not discuss the impact of topology on gradients.

Recently, random graph concepts have been used in deep learning. For instance,~\cite{randwire, dnw} utilize standard random graphs such as Barabasi-Albert (BA)~\cite{bara} or Watts-Strogatz (WS)~\cite{swnets} models for NAS. However, like other NAS research,~\cite{randwire, dnw} do \textit{not} connect the topology with the gradient flow. In contrast, by considering the concatenation-type skip connections, we aim to quantify the link between topology, gradients, and accuracy. Further, while we do build new models as a proof-of-concept of our theoretically grounded metric, we do not conduct any NAS. Conducting full NAS guided by theoretical metrics is left as a future work.
\begin{figure*}[t]
\centering
\includegraphics[width=1\textwidth]{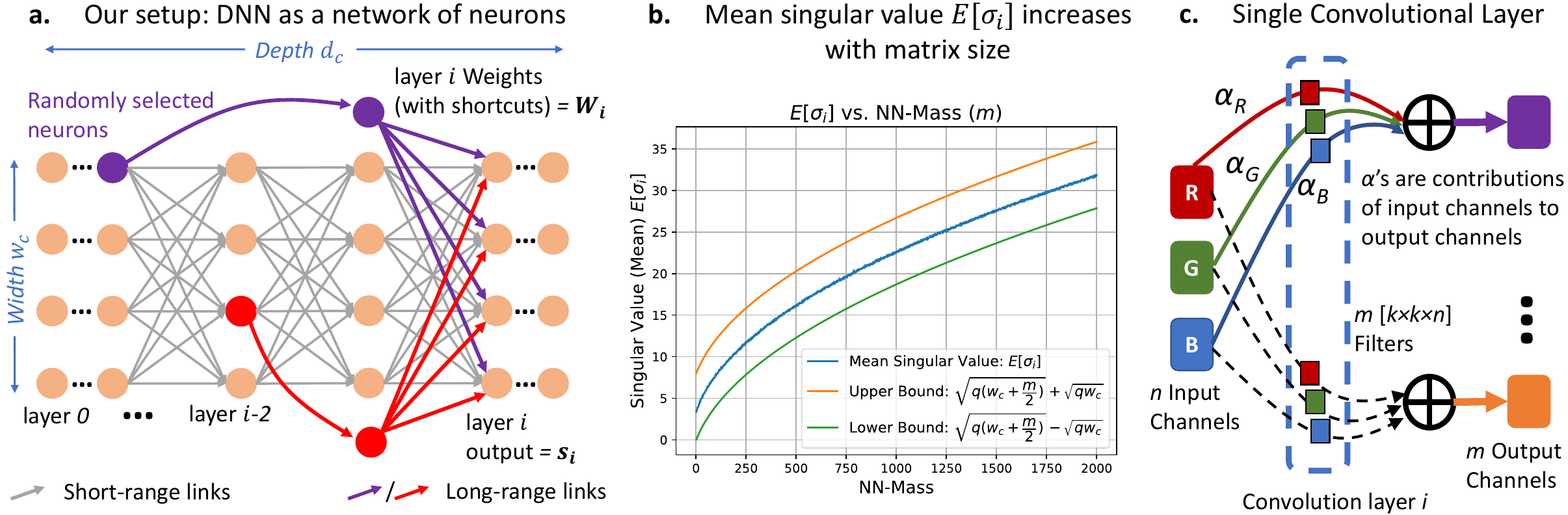}\vspace{-2.5mm}
	\caption{(a) Setup: DNN (depth $d_c$, width $w_c$) has layer-by-layer connections due to MLP (gray links) with random concatenation-type skip connections (purple/red links).  (b)~Simulation of Gaussian matrices sized $(w_c+m/2, w_c)$, where $w_c=16$: Mean singular values increase as NN-Mass ($m$) increases and is bounded as shown in Proposition~\ref{prop2} (Appendix~\ref{prProp2App}). (c)~Convolutional layers form a similar topological structure as MLP: All input channels contribute to all output channels.\vspace{-3mm}}
\label{fig2}
\end{figure*}

\vspace{-2mm}

\paragraph{Preliminaries.} We use the following established concepts:\vspace{-1mm}
\begin{definition}[Average Degree~\cite{networksci}]
Average degree ($\hat{k}$) of a network represents the average number of connections across all nodes, $\hat{k} = \text{\#edges}/\text{\#nodes}$.\vspace{-1mm}
\end{definition}

Average degree and degree distribution (\textit{i.e.}, distribution of nodes' degrees) are important topological characteristics which directly affect how information flows through a network. 
How fast a signal can propagate through a network heavily depends on the network topology. 

\begin{definition}[Layerwise Dynamical Isometry (LDI)~\cite{sigprop}]
A deep network satisfies LDI if the singular values of Jacobians at initialization are close to 1 for all layers. Specifically, for a multilayer feed-forward network, let $\bm{s_i}$ ($\bm{W_i}$) be the output (weights) of layer $i$ such that $\bm{s_i} =\phi(\bm{h_i}), \bm{h_i}=\bm{W_i s_{i-1}} + \bm{b_i}$; then, the Jacobian matrix at layer $i$ is defined as: $\lj = \frac{\partial \bm{s_{i}}}{\partial \bm{s_{i-1}}} = \bm{D_i W_i}$. Here, $\lj \in \mathbb{R}^{w_i, w_{i-1}}$, $w_i$ is the number of neurons in layer $i$. $\bm{D^{jk}_i}=\phi'(\bm{h_i})\delta_{jk}$. $\phi'$ denotes the derivative of non-linearity $\phi$ and $\delta_{jk}$ is Kronecker delta. Then, if the singular values $\sigma_j$ for all $\lj$ are close to 1, then the network satisfies the LDI.
\label{def2}
\end{definition}

\vspace{-1mm}
LDI discourages the signal propagating through the DNN from getting attenuated or amplified too much; this ensures faithful propagation of gradients~\cite{saxe2013exact}. 

\section{Topological Properties of DNNs}\label{app}
We first describe our setup of DNNs with DenseNet-type skip connections and propose the new topological metrics. We then demonstrate the theoretical relationship between the topology and gradient propagation.

\subsection{Modeling DenseNet-type Skip Connections}\label{mod}
We start with a generic MLP setup with $d_c$ layers containing $w_c$ neurons each and assume DenseNet-type skip connections superimposed on top of a typical MLP structure (see Fig.~\ref{fig2}(a)). Henceforth, unless stated otherwise, DenseNet-type skip connections will be simply referred to as skip connections. Specifically, all neurons at layer $i$ receive skip connections from a maximum of $t_c$ neurons from previous layers. That is, we \textit{randomly} select min$\{w_c(i-1), t_c\}$ neurons from layers $0,1,\ldots,(i-2)$, and concatenate them at layer $i-1$ (see Fig.~\ref{fig2}(a))\footnote{Here, $w_c(i-1)$ is the total number of candidate neurons from layers $0,1,\ldots,(i-2)$ that can supply skip connections; if the \textit{maximum} number of neurons $t_c$ that can supply skip connections to the current layer exceeds total number of possible candidates, then all neurons from layers $0,1,\ldots,(i-2)$ are selected. Neurons are concatenated similar to how channels are concatenated in DenseNets~\cite{densenet}.}; the concatenated neurons then pass through a fully-connected layer to generate the output of layer $i$ ($\bm{s_i}$). As a result, the weight matrix $\bm{W_i}$ (which is used to generate $\bm{s_i}$) gets additional weights to account for the incoming skip connections. Similar to recent NAS research~\cite{ameetNAS}, we select links randomly because random architectures are often as competitive as the carefully designed models. 
Moreover, the random skip connections on top of fixed short-range links make our architectures a small-world network (Fig.~\ref{swa}, Appendix~\ref{swaApp})~\cite{swnets} which allows us to use graph/network concepts to study their topology.

An important advantage of the above setup is that we can control the density of skip connections (using $t_c$) to study the topological properties over many DNNs. If the skip connections encompass all-to-all connections, this will result in the original DenseNet architecture. Like standard CNNs (Resnets/DenseNets), we can generalize the setup to contain multiple ($N_c$) cells of width $w_c$, depth $d_c$; skip connections exist only \textit{within} a cell and not across cells.

\subsection{Proposed Metrics}\label{md}
Our key objective is to quantify what topological characteristics of DNNs with DenseNet-type skip connections affect their accuracy and gradient flow. We then exploit such properties to \textit{directly} design efficient CNNs by looking at such properties. To this end, we propose two new metrics called \textit{Cell-Density} and \textit{NN-Mass}, as defined below. 
\begin{definition}[Cell-Density]\label{cellDensityDef}
Density of a cell quantifies how densely its neurons are connected via skip connections. Formally, for a cell $c$, cell-density $\rho_c$ is given by:
\begin{equation}
\begin{aligned}
\rho_c &= \frac{Actual \ \text{\#skip connections within cell c}}{\text{Total possible \#skip connections within cell c}}\\
       &= \frac{2\sum_{i=2}^{d_c-1}\text{min}\{w_c (i-1), t_c\}}{w_c (d_c - 1) (d_c - 2)}
\end{aligned}
\label{rho1}
\end{equation}
\end{definition}

For complete derivation, please refer to Appendix~\ref{derNND}. 
Informally, density is basically \textit{mass/volume}. Let \textit{volume} be the total number of neurons in a cell ($w_c\times d_c$). Then, we define the NN-Mass ($\nnMass$) as follows:

\vspace{2mm}
\begin{definition}[Mass of DNNs]\label{nnMassDef} NN-Mass is defined as the sum (over all cells) of product of Cell-Density ($\rho_c$) and number of neurons per cell.
\begin{equation}
	\begin{aligned}
 		\nnMass &= \sum_{c=1}^{N_c} w_c d_c \rho_c= \sum_{c=1}^{N_c} \frac{2 d_c\sum_{i=2}^{d_c-1}\text{min}\{w_c (i-1), t_c\}}{(d_c - 1) (d_c - 2)}
	\end{aligned}
\label{mass1}
\end{equation}
\end{definition}

As explained in the next section, NN-Mass quantifies how effectively information can flow through a given DNN topology. For a given width ($w_c$), models with similar NN-Mass, but different depths ($d_c$) and \#Params, should exhibit a similar gradient flow and, thus, achieve a similar accuracy. Note that, NN-Mass is a function of network width, depth, and skip connections (\textit{i.e.}, the topology of the network). For a fixed number of cells, an architecture can be completely specified by \textit{$\{$depth, width, maximum skip connection candidates$\}$} per cell = $\{d_c, w_c, t_c\}$. Hence, to create different architectures with DenseNet-type skip connections, we vary $\{d_c, w_c, t_c\}$ to create architectures with random \#Params/FLOPS/layers, and NN-Mass. We then train these architectures and characterize their accuracy, topology, and gradient propagation to understand the relationships among them. But first, we provide our theoretical analysis.

\subsection{Relationships among topology, NN-Mass and gradient propagation}\label{theoMass}

Without loss of generality, we assume that the DNN (same setup as above) has only one cell of width $w_c$ and depth $d_c$.

\setcounter{proposition}{0}

\begin{proposition}[\textbf{NN-Mass and average degree}]
The average degree of a DenseNet-type deep network with NN-Mass $\nnMass$ is given by $\hat{k} = w_c+\nnMass/2$.
\label{prop1}
\end{proposition}

\vspace{-2mm}
\noindent
The proof of the above result is given in Appendix~\ref{prProp1App}.

\noindent
\textbf{Intuition.}
Proposition~\ref{prop1} states that the average degree of a deep network is $w_c + \nnMass/2$, which, given the NN-Mass $\nnMass$, is independent of depth $d_c$. The average degree indicates how well-connected the network is. Hence, it controls how effectively the information can flow through a given topology. Therefore, for a given width and NN-Mass, the average amount of information that can flow through various architectures (with different \#Params/layers) should be similar (due to the same average degree). Thus, we hypothesize that these topological characteristics might constrain the amount of information being learned by DNNs. Next, we show the impact of topology on gradient propagation.

\begin{proposition}[\textbf{NN-Mass and LDI}]
Consider the case of deep linear networks with concatenation-type skip connections, where each layer is initialized using independently and identically distributed values with initialization variance $q$. For this setup, suppose we are given a small network $f_S$ (depth $d_S$) and a large network $f_L$ (depth $d_L$, $d_L>>d_S$), both with same initialization scheme, NN-Mass $\nnMass$, and width $w_c$. Then, the mean singular value of the initial layerwise Jacobian ($\mathbb{E}[\sigma]$) for both networks is bounded as follows: \vspace{-2mm}
\begin{equation*}
    \sqrt{q(w_c+m/2)}-\sqrt{qw_c} \leq \mathbb{E}[\sigma] \leq \sqrt{q(w_c+m/2)}+\sqrt{qw_c}\vspace{-2mm}
\label{prop2Bound}
\end{equation*}
That is, the LDI for both models does not depend on the depth if the initialization variance ($q$) for each layer is depth-independent (which is the case for many initialization schemes). Hence, for such networks, models with similar width and NN-Mass result in similar gradient properties, even if their depths and \#Params are different.
\label{prop2}
\end{proposition}

\vspace{-1mm}
\noindent
\textit{Proof:} A formal proof of the above result and the bounds under the deep linear network~\cite{saxe2013exact,linear_network1,linear_network2,linear_network3} assumption is given in Appendix~\ref{prProp2App}. The discussion below is more informal and explains how the above result works for both linear and non-linear DNNs with DenseNet-type skip connections.

To prove this result, it suffices to show that the initial Jacobians $\lj$ have similar properties for both models (and thus their singular values are similar). For our setup, the output of layer $i$, $\bm{s_{i}} = \phi(\bm{W_{i}x_{i-1}} + \bm{b_i})$, where $\bm{x_{i-1}} = \bm{s_{i-1}} \cup \bm{y_{0:i-2}}$ concatenates output of layer $i-1$ ($\bm{s_{i-1}}$) with the neurons $\bm{y_{0:i-2}}$ supplying the skip connections (random $\text{min}\{w_c(i-1), t_c\}$ neurons selected uniformly from layers $0$ to $i-2$). Hence, $\lj = \partial \bm{s_{i}}/\partial \bm{x_{i-1}} = \bm{D_i  W_i}$. Compared to a typical MLP (see Definition~\ref{def2}), the sizes of $\bm{D_i}$ and $\bm{W_i}$ increase to account for incoming skip connections.

For two models $f_S$ and $f_L$, the layerwise Jacobian ($\lj$) can have two kinds of properties: (\textit{i})~The distribution of values inside Jacobian matrix for $f_S$ and $f_L$ can be different, and/or (\textit{ii})~The sizes of layerwise Jacobian matrices for $f_S$ and $f_L$ can be different. Hence, our objective is to show that when the width ($w_c$) and NN-Mass ($m$) are similar, irrespective of the depth of the model (and thus irrespective of \#Params/FLOPS), both the distribution and the size of initial layerwise Jacobians are similar.

Let us start by considering a linear network: in this case, $\lj = \bm{W_i}$. Since the LDI looks at the properties of layerwise Jacobians \textit{at initialization}, and because all models are initialized the same way (\textit{e.g.}, Gaussians with variance scaling\footnote{Variance scaling methods also take into account the number of input/output units. Hence, if the width is the same between models of different depths, the distribution at initialization is still similar.}), the values inside $\lj$ for both $f_S$ and $f_L$ have same distribution (\textit{i.e.}, point (\textit{i}) above is satisfied). We next show that even the sizes of layerwise Jacobians for both models are similar if the width and NN-Mass are similar. 

How is topology related to the layerwise Jacobians? Since the average degree is same for both models (see Proposition~\ref{prop1}), on average, the number of incoming skip connections at a typical layer is $w_c \times \nnMass/2$. In other words, since the degree distribution for the random skip connections is Poisson~\cite{randomDegDist} with average degree $\rgKbar\approx \nnMass/2$ (see Eq.~\ref{krg1}, Appendix~\ref{prProp1App}), an average $\nnMass/2$ neurons supply skip connections to each layer\footnote{Poisson process assumes a constant rate of arrival of skip connections.}. Therefore, the Jacobians will theoretically have the same dimensions $(w_c+\nnMass/2, w_c)$ irrespective of the depth of the neural network (\textit{i.e.}, point (\textit{ii}) is also satisfied). 

So far, the discussion has considered only a linear network. For a non-linear network, the Jacobian is given as $\lj = \bm{D_i W_i}$. As explained in~\cite{sigprop}, $\bm{D_i}$ depends on pre-activations $\bm{h_i} = \bm{W_i x_{i-1}}+ \bm{b_i}$. As established in several deep network mean field theory studies~\cite{poole2016exponential, tarnowski2018dynamical}, the distribution of pre-activations at layer $i$ ($\bm{h_i}$) is a Gaussian $\mathcal{N}(0, q_i)$ due to the central limit theorem. Similar to~\cite{sigprop, pennington2017resurrecting}, if the input $\bm{h_0}$ is chosen to satisfy a fixed point $q_i = q^*$, the distribution of $D_i$ becomes independent of the depth ($\mathcal{N}(0,q^*)$). Therefore, the distribution of both $\bm{D_i}$ and $\bm{W_i}$ is similar for different models irrespective of the depth, even for non-linear networks. Moreover, the sizes of the matrices will be similar due to similar average degree in $f_S$ and $f_L$.

Hence, the size and distribution of values in the Jacobian matrix are similar for both the large and the small model (provided the width and NN-Mass are similar); that is, the distribution and mean singular values will also be similar. Thus, LDI is equivalent between different depth DNNs if their width and NN-Mass are similar. As a result, such models have similar gradient flow properties. \qed

To verify the bounds provided in Proposition~\ref{prop2}, we numerically simulate the mean singular values of layerwise Jacobians for deep linear networks using standard Gaussian ($q=1$) matrices of sizes ($w_c+\nnMass/2, w_c$). Specifically, we vary $\nnMass$ for a given width $w_c$ and see the impact of this size variation on mean singular values. Fig.~\ref{fig2}(b) shows that as NN-Mass varies, the mean singular values increase and lie within the bounds of Proposition~\ref{prop2}. Note that, our results should \textit{not} be interpreted as bigger models yield larger mean singular values. We show in the next section that the relationship between the \#Params and mean singular values is significantly worse than that for NN-Mass. Hence, it is the topological properties that enable LDI in different deep networks and \textit{not} the \#Params.
\vspace{-3mm}
\paragraph{Remark 1 (NN-Mass formulation is same for DenseNet-type CNNs).} 
Fig.~\ref{fig2}(c) shows a typical convolutional layer. Since all channel-wise convolutions are added together, each output channel is some function of all input channels. This makes the topology of CNNs similar to that of our MLP setup. The key difference is that the nodes in the network (see Fig.~\ref{fig2}(a)) are now channels and not individual neurons. Of note, for our CNN setup, we use three cells (similar to DenseNets). More details on CNN setup (including a concrete example for NN-Mass calculations) are given in Appendices~\ref{cnnSetupApp} and~\ref{egApp}.

\begin{figure}[t]
\begin{center}
  \includegraphics[width=0.45\textwidth]{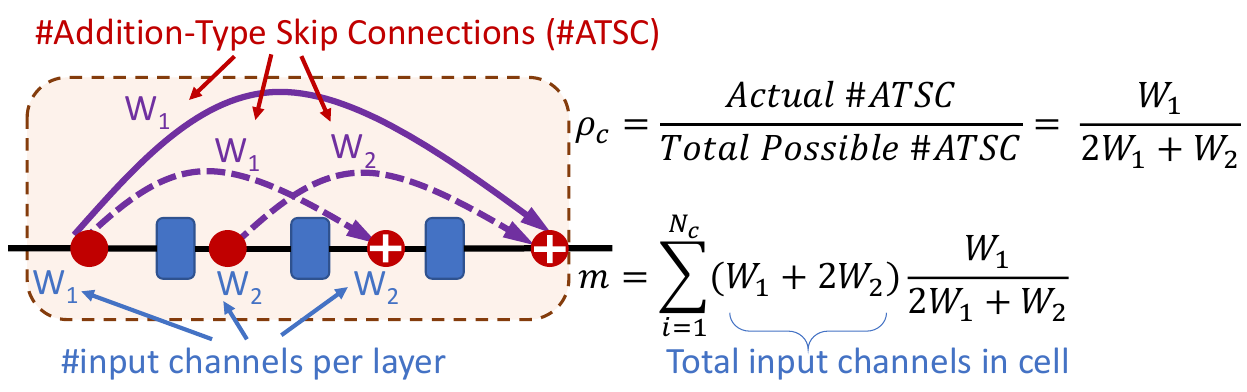}
  \vspace{-3mm}
  \caption{NN-Mass for bottleneck ResNets/MobileNets. Dotted purple lines are possible ATSC (not actually present). Solid purple ATSC are present in MobileNets/ResNets.\vspace{-4mm}}
  
\label{fig:ResNet}
\end{center}
\end{figure}

\vspace{-3mm}
\paragraph{Remark 2 (NN-Mass generalizes to ResNets and MobileNets).} 

Note that, ResNets/MobileNet-v2 have Addition-Type Skip Connections (ATSC). Then, following Definitions~\ref{cellDensityDef} and~\ref{nnMassDef}, cell-density ($\rho_c$) and NN-Mass for ResNets/MobileNets are defined as: \vspace{-2mm}
\begin{equation}
	    \rho_c=\frac{\text{Actual \#ATSC}}{\text{Total possible \#ATSC}}, \ \ \ m = \sum\nolimits_{N_c} i_c \times \rho_c\vspace{-2mm}
\label{genMass}
\end{equation}
where, $i_c$ is the total input channels within one cell. For instance, for the bottleneck cells shown in Fig.~\ref{fig:ResNet}, $i_c=W_1+2W_2$, where $W_1$ and $W_2$ are number of input channels at various layers within the bottleneck cell. Due to one-to-one, channel-wise additions, the actual \#ATSC = $W_1$ since a maximum of $W_1$ channels can be added (this cannot exceed the \#input channels at the source layer; see solid purple line in Fig.~\ref{fig:ResNet}). In bottleneck cells, ATSC can be present at two other locations (see dotted purple lines in Fig.~\ref{fig:ResNet}), each can supply $W_1$ and $W_2$ links, respectively. Hence, $\rho_c$ and NN-Mass can be computed as shown in Fig.~\ref{fig:ResNet} (right) using Eq.~\eqref{genMass}. Equations shown in Fig.~\ref{fig:ResNet} work for both ResNet and MobileNet bottleneck cells, and a similar process follows for the ResNet-Basicblock cell.

\begin{figure*}[tb]
\centering
    \includegraphics[width=1.0\textwidth]{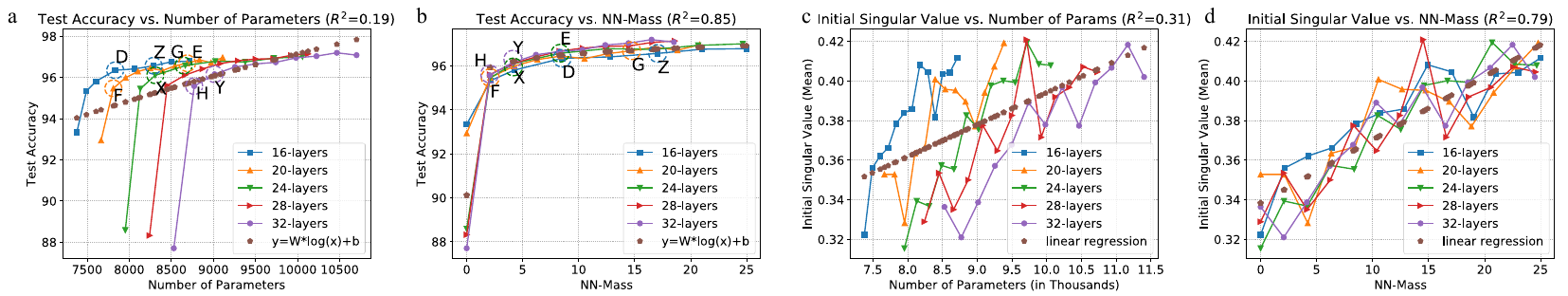}
\vspace{-10pt}
\centering
\vspace{-10pt}
\caption{MNIST results: (a)~Models with different \#Params achieve similar test accuracy. (b)~Test accuracy curves of models with different depths/\#Params concentrate when plotted against NN-Mass (test accuracy std. dev. $\sim$ $0.05-0.34\%$). (c,d)~Mean singular values of $\lj$ are much better correlated with NN-Mass ($\bm{R^2=0.79}$) than with \#Params ($R^2=0.31$).\vspace{-3mm}}
\label{mlp_mnist}
\end{figure*}
\begin{figure*}[tb]
    \centering
    \includegraphics[width=0.85\textwidth]{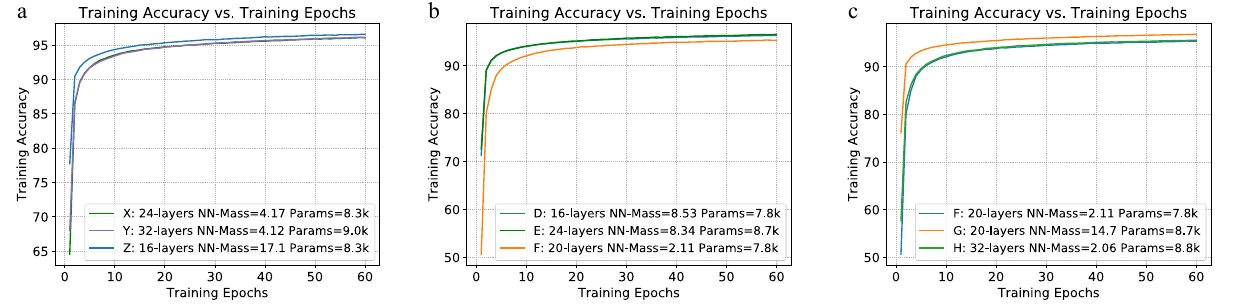}
    \vspace{-10pt}
    \caption{Models X and Y have the same NN-Mass and achieve very similar training convergence, even though they have highly different \#Params and depth. Model Z has significantly fewer layers than Y but the same \#Params, yet achieves a faster training convergence than Y (Z has higher NN-Mass than Y). The above conclusions hold true for models (D, E, and F) and (F, G and H). Note that, training convergence curves are nearly coinciding for models with same NN-Mass.\vspace{-15pt}}
    \label{mlp_training}
\end{figure*}

We next provide extensive empirical evidence for our theoretical insights on topology, gradient propagation, LDI, and model accuracy (Proposition~\ref{prop2}).
\section{Experimental Setup and Results}
\label{exp}

\subsection{Experimental Setup}
For experiments on MLPs and CNNs, we generate random architectures (within our setup of DenseNet-type skip connections) with different NN-Mass and number of parameters by varying $\{d_c, w_c, t_c\}$. For random MLPs with different $\{d_c, t_c\}$ and $w_c=8$ (\#cells = 1), we conduct the following experiments on the MNIST dataset: 
(\textit{i})~We explore the impact of varying \#Params and NN-Mass on the test accuracy; 
(\textit{ii})~We demonstrate how LDI depends on NN-Mass and \#Params; 
(\textit{iii})~We further show that models with similar NN-Mass (and width) result in similar training convergence, despite having different depths and \#Params.

After the extensive empirical evidence for our theoretical insights (\textit{i.e.}, the connection between gradient propagation and topology), we next move on to CNN architectures. We conduct the following experiments: (\textit{i})~For three-cell CNNs with random concatenation-type skip connections (\textit{i.e.}, the DenseNet setup), we show that NN-Mass can identify CNNs that achieve similar test accuracy, despite having highly different \#Params/FLOPS/layers; (\textit{ii})~We show that NN-Mass is a significantly more effective indicator of model performance than parameter counts; (\textit{iii})~For DenseNet setup, we perform the above experiments for CIFAR-10, CIFAR-100, and ImageNet datasets; (\textit{iv})~We further demonstrate that NN-Mass works for standard ResNets, Wide-ResNets, and MobileNets on ImageNet.

Finally, we exploit NN-Mass to directly design efficient DenseNet-type CNNs (for CIFAR-10) and efficient MobileNet-like networks (for ImageNet) which achieve accuracy comparable to significantly larger models. Overall, we train hundreds of different MLP and CNN architectures with each MLP (CNN) repeated five (three) times with different random seeds, to obtain our results. More setup details (\textit{e.g.}, architecture details, learning rates, \textit{etc.}) are given in Appendix~\ref{appExpDetails} (see Tables~\ref{expAll},~\ref{expAll1}, and~\ref{expAll3}).

\subsection{MLP Results (MNIST/Synthetic Data): \\Topology vs. Gradient Propagation}\label{mlpRes}

\vspace{-2mm}
\paragraph{Test Accuracy.}
Fig.~\ref{mlp_mnist}(a) shows test accuracy \textit{vs.} \#Params of DNNs with different depths on the MNIST dataset. As evident, even though many models have different \#Params, they achieve a similar test accuracy. On the other hand, when the same set of models are plotted against NN-Mass, their test accuracy curves cluster together tightly, as shown in Fig.~\ref{mlp_mnist}(b). To further quantify the above observation, we generate a linear fit between test accuracy \textit{vs.} log(\#Params) and log(NN-Mass) (see brown markers in Fig.~\ref{mlp_mnist}(a,b)). For NN-Mass, we achieve a significantly higher goodness-of-fit $R^2=0.85$ than that for \#Params ($R^2=0.19$). This demonstrates that NN-Mass can identify DNNs that achieve similar accuracy, even if they have a highly different number of parameters/FLOPS\footnote{For our DenseNet-based setup, more parameters lead to more FLOPS. Results for FLOPS are given for CNNs in Appendix~\ref{flopsApp}.}/layers. We next investigate the gradient propagation properties to explain the test accuracy results.

\vspace{-2mm}
\paragraph{Layerwise Dynamical Isometry (LDI).}
We calculate the mean singular values of initial layerwise Jacobians, and plot them against \#Params (see Fig.~\ref{mlp_mnist}(c)) and NN-Mass (see Fig.~\ref{mlp_mnist}(d)). Clearly, NN-Mass ($R^2=0.79$) is far better correlated with the mean singular values than \#Params ($R^2=0.31$). More importantly, just as Proposition~\ref{prop2} predicts, these results show that models with similar NN-Mass and width have equivalent LDI properties, irrespective of the total depth (and, thus \#Params) of the network. For example, even though the 32-layer models have more parameters, they have similar mean singular values as the 16-layer DNNs. This clearly suggests that the gradient propagation properties are heavily influenced by the topological characteristics like NN-Mass, and not just by DNN depth and \#Params.

\vspace{-2mm}
\paragraph{Training Convergence.}
The above results suggest the following hypotheses: (\textit{i})~If the gradient flow between DNNs (with similar NN-Mass and width) is similar, their training convergence should be similar, even if they have highly different \#Params and depths; (\textit{ii}) If two models have same \#Params (and width), but different depths and NN-Mass, then the DNN 
with higher NN-Mass should have faster training convergence (since its mean singular value will be higher -- see the trend in Fig.~\ref{mlp_mnist}(d)).

To demonstrate that both hypotheses above hold true, we pick three groups of three models each --  (X, Y, and Z), (D, E, and F), and (F, G, and H) from Fig.~\ref{mlp_mnist}(a,b) and plot their training accuracy \textit{vs.} \#epochs in Fig.~\ref{mlp_training}. In Fig.~\ref{mlp_training}(a), Models X and Y have similar NN-Mass but Y has more \#Params and depth than X. Model Z has far fewer layers and nearly the same \#Params as X, but has higher NN-Mass. Fig.~\ref{mlp_training}(a) shows the training convergence results for X, Y, and Z. As evident, the training convergence of model X (8.3K Params, 24-layers) nearly coincides with that of model Y (9.0K Params, 32-layers). Moreover, even though model Z (8.3K Params, 16-layers) is shallower than the 32-layer model Y (and has far fewer \#Params), training convergence of Z is significantly faster than that of Y (due to higher NN-Mass and, therefore, better LDI). These results clearly show the evidence supporting Proposition \ref{prop2}, and emphasize the concrete links among topology, gradient propagation and model performance for DNNs with DenseNet-type skip connections. Similar observations are found for models (D, E, and F) and (F, G, and H) as shown in Fig.~\ref{mlp_training}(b,c).

\subsection{CNN Results on CIFAR-10/100 and ImageNet}\label{naseRes}
Having established a concrete relationship between gradient propagation and topological properties, we now show that NN-Mass can identify efficient CNNs that achieve similar accuracy as models with significantly higher \#Params/FLOPS/layers. Unless specified, our CNN models belong to the DenseNet setup. We will explicitly indicate when the results are for ResNets/WRNs/MobileNets.

\begin{figure}[htb]
    \centering
    \includegraphics[width=0.45\textwidth]{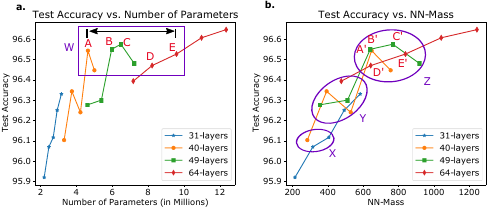} \vspace{-2mm}
    \caption{CIFAR-10 Width Multiplier $wm=2$: (a)~Models with very different \#Params (box W) achieve similar test accuracies. (b)~Models with similar accuracy often have similar NN-Mass: Models in W cluster into Z. Results are reported as the mean of three runs (std. dev. $\sim$ $0.1\%$).\vspace{-2mm}}
    \label{avpavm2}
\end{figure}

\vspace{-2mm}
\paragraph{Model Performance for CIFAR-10 dataset.} 
Fig.~\ref{avpavm2}(a) shows the test accuracy of various CNNs \textit{vs.} total \#Params. As evident, models with highly different \#Params (\textit{e.g.}, see models A-E in box W), achieve a similar test accuracy. Note that, there is a large gap in the model size: CNNs in box W 
range from 5M parameters (model A) to 9M parameters (models D,E). Again, as shown in Fig.~\ref{avpavm2}(b), when plotted against NN-Mass, the test accuracy curves of CNNs with different depths cluster together (\textit{e.g.}, models A-E in box W cluster into A'-E' within bucket Z). Hence, NN-Mass identifies CNNs with similar accuracy, despite having highly different \#Params/layers. The same holds true for models within X and Y boxes. More results with different width multipliers are given in Appendix~\ref{diffWC10_App}. For higher width values, the models tend to cluster even more tightly for NN-Mass.
\begin{figure*}[htb]
\centering\vspace{-5mm}
\includegraphics[width=0.96\textwidth]{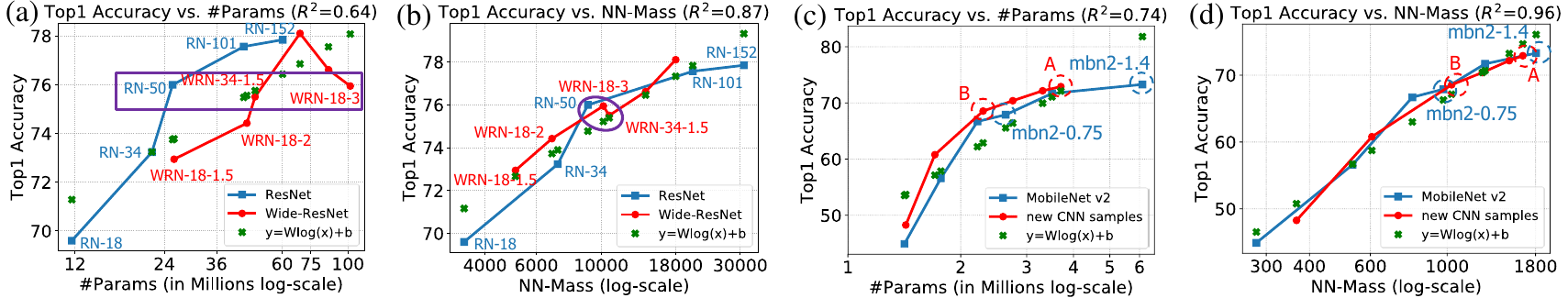}\vspace{-2.5mm}
    \caption{ImageNet results: (a,b) Top1 accuracy vs. (a) \#Params ($R^2$=$0.64$);  (b) NN-Mass ($\bm{R^2}$\textbf{=}$\bm{0.87}$) for standard ResNet and Wide-ResNets. (c,d) Top1 accuracy vs. (c) \#Params ($R^2$=$0.74$); (d) NN-Mass ($\bm{R^2}$\textbf{=}$\bm{0.96}$) for MobileNet-v2 and some newly sampled CNNs (Better networks are highlighted with \textit{red} letters.).}\vspace{0mm}
	
\label{fig:atsc}
\vspace{-5mm}
\end{figure*}

\vspace{-3mm}
\paragraph{NN-Mass vs. Parameter Count.} As shown in Fig.~\ref{genP} in Appendix~\ref{appResComp}, for $wm=2$, \#Params yield an $R^2=0.76$ which is lower than that for NN-Mass ($R^2=0.84$, see Fig.~\ref{genP}(a, b)). However, for higher widths ($wm=3$), the parameter count completely fails to predict accuracy ($R^2=0.14$ in Fig.~\ref{genP}(c)). For the same width, NN-Mass achieves a significantly higher $R^2=0.90$ (see Fig.~\ref{genP}(d)).

\vspace{-3mm}
\paragraph{Results for CIFAR-100 Dataset.}
We now corroborate our main findings on CIFAR-100 dataset which is significantly more complex than CIFAR-10. To this end, we train the models in Fig.~\ref{avpavm2} on CIFAR-100. Fig.~\ref{avpavm2c100} (see Appendix~\ref{c100App}) once again shows that several models with highly different number of parameters achieve similar accuracy. Moreover, Fig.~\ref{avpavm2c100}(b) demonstrates that these models get clustered when plotted against NN-Mass. Further, a high $R^2=0.84$ is achieved for a linear fit on the accuracy vs. log(NN-Mass) plot (see Appendix~\ref{c100App} and Fig.~\ref{avpavm2c100}).

\vspace{-3mm}
\paragraph{DenseNet-type CNNs for ImageNet and other results.} We provide the ImageNet results in Appendix \ref{dense_imagenet}. More results for the depthwise convolutions (DSConv) with concatenation-type skip links for CIFAR-10 are given in Appendix \ref{dense_ds_conv}. Again, NN-Mass identifies CNNs with similar accuracy while having significantly different \#Params.

\vspace{-3mm}
\paragraph{Results for \#FLOPS.}
The results for \#FLOPS follow a very similar pattern as \#Params (see Fig.~\ref{flops} in Appendix~\ref{flopsApp}). In summary, we show that NN-Mass can identify models that yield similar test accuracy, despite having very different \#Params/FLOPS/layers. 

\vspace{-3mm}
\paragraph{NN-Mass on Standard DenseNets and VGG.} So far, we have used a generalized version of DenseNets which allows us to vary the density of skip connections. For standard DenseNets~\cite{densenet}, cell-density (Eq.~\ref{rho1}) = 1 (all-to-all connections); thus, NN-Mass for DenseNet = $\sum_{all\ cells}$[(\#channels per layer for this cell) $\times$ (\#layers per cell)]. For VGG-like models, there are no shortcuts, so NN-Mass = 0. Our theory works for NN-Mass = 0: Fig.~\ref{mlp_mnist}(b) shows two clusters for [low-depth (16,20) NN-Mass 0] models and [high-depth (24,28,32) NN-Mass 0] models. This is \textit{not} surprising: without shortcuts, the gradient diminishes as depth increases (e.g., see Resnets~\cite{resnet}). Same holds for our NN-Mass=0 CNNs on ImageNet and CIFAR-10. 

\vspace{-4mm}
\paragraph{NN-Mass works for ResNets and MobileNets.} For ResNets and MobileNets, we calculate the NN-Mass values using Remark~2. Wide-ResNet (WRN) paper\footnote{We directly use the accuracy/\#Params from Tables 7,8 in the Wide-ResNet paper~\cite{wrn} (no training is performed).} provides \#Params and test accuracy for standard ResNets and WRNs for ImageNet (Fig.~\ref{fig:atsc}(a)). Fig.~\ref{fig:atsc}(b) shows Top1 accuracy vs. NN-Mass. Clearly, NN-Mass outperforms \#Params for predicting accuracy ($\bm{R^2}$\textbf{=}$\bm{0.87}$ vs.  $R^2$=$0.64$). Note that, RN-50, WRN-34-1.5, and WRN18-3 (26M-101M \#Params) achieve similar accuracy (purple box in Fig.~\ref{fig:atsc}(a)), and cluster together on NN-Mass plot (purple circle in Fig.~\ref{fig:atsc}(b)). 

Fig.~\ref{fig:atsc}(c,d) shows the Top1 accuracy of MobileNet-v2 (mbn2) vs. \#Params (Fig.~\ref{fig:atsc}(c)) and NN-Mass (Fig.~\ref{fig:atsc}(d)) on ImageNet. The blue line shows the standard MobileNet-v2 models ($N_c$=$17$; \{0.75, 1.4\} are width-multipliers). The red line shows \textit{new} CNNs sampled by changing width multipliers and total depth ($N_c$=$22$).  
Again, NN-Mass significantly outperforms \#Params ($\bm{R^2}$\textbf{=}$\bm{0.96}$ vs. $R^2$=$0.74$). 
Hence, NN-Mass works for standard ResNets/WRNs/MobileNet-v2.

\vspace{-3mm}
\paragraph{Directly designing compressed CNNs with NN-Mass.} We first design regular DenseNet-type CNNs (no DSConv) with skip connections for CIFAR-10 dataset and show how such models can be compressed directly using NN-Mass equation~(\ref{mass1}) without searching for an efficient network. Detailed procedure is provided in Appendix \ref{appDirRes}. 
To summarize: (\textit{i})~We generate candidate CNNs by varying \{$d_c$,$w_c$,$t_c$\} (for DenseNets) or \{$N_c$, width-multiplier\} (for MobileNets). Note, we do \textit{not} train these CNNs. (\textit{ii})~Next, find the compressed CNN with highest NN-Mass (using Definition \ref{nnMassDef}) given the \#Params/MACs constraints. (\textit{iii})~Train this model to verify its accuracy. This is our compressed model. 

As shown in Table~\ref{massModC}, our models reach a test accuracy of $96.82\%$-$97.00\%$ on CIFAR-10, while reducing the number of parameters and FLOPS by up to $3\times$ over large CNNs (e.g., 3.82M \textit{vs.} 11.89M parameters). As a reference, DARTS~\cite{darts}, a competitive NAS baseline, achieves a comparable ($97\%$) accuracy with slightly lower 3.3M parameters. Note that, our objective is \textit{not} to beat DARTS or any other baseline, but rather to provide theoretical insights into the behavior of DNNs with DenseNet-type skip connections. The DARTS datapoint is chosen just to show that our efficient, high-accuracy, theoretically grounded CNNs (that do not use specialized search spaces like NAS) are capable of reaching state-of-the-art accuracy and, hence, are practically useful. 
\begin{table}[tb]
\caption{Exploiting NN-Mass for Model Compression on CIFAR-10 Dataset. All results are reported as mean~$\pm$~standard deviation of three runs. DARTS results are from~\cite{darts}.\vspace{-6mm}}
\label{massModC}
\begin{center}
\scalebox{0.73}{
\begin{tabular}{c|c|c|c|c}
    \hline
    Model & \#Params/\#FLOPS & \#layers &NN-Mass & Test Accuracy \\
    \hline
    DARTS$^{\text{I}}$ (NAS) & 3.3M/- & - &- & 97\% \\
    \hline
    DARTS$^{\text{II}}$ (NAS) & 3.3M/- & - &- & 97.24\% \\
    \hline
    \hline
    Manual model&  11.89M/3.63G & 64 & 1126 & 97.02\% \\
    \hline
    Manual model &  8.15M/2.54G& 64& 622 &96.99\% \\
    \hline
    \hline
    NN-Mass based  &5.02M/1.59G  &40  &755 &97.00\% \\
    \hline
    NN-Mass based  &4.69M/1.51G  &37  &813 &96.93\% \\ 
    \hline
    NN-Mass based  &3.82M/1.2G  &31  &856 &96.82\% \\
    \hline
\end{tabular}
}\vspace{-3mm}
\end{center}
{\scriptsize I -- DARTS First Order, II -- DARTS Second Order}\vspace{-4mm}
\end{table}

\begin{table}[h]
    \centering
    \caption{Compressed MobileNets via NN-Mass on ImageNet}
    \scalebox{0.73}{
    \begin{tabular}{c|c|c|c|c}
    \hline
         Network & NN-Mass & Top1 & \#Params & MACs   \\
    \hline \hline
         MobileNetV2 (1.4)& 1807 & 73.3& 6.1M& 601M \\
    \hline
         \textbf{NN-Mass based A (Fig.~\ref{fig:atsc})} &  1654& 72.9 & \textbf{3.7M}& \textbf{393M }\\
    \hline \hline
         MobileNetV2 (0.75)& 975.0 & 67.9& 2.6M& 220M \\
    \hline
         \textbf{NN-Mass based B (Fig.~\ref{fig:atsc})} & 1030 & 68.56 & \textbf{2.3M}& \textbf{200M} \\
    \hline
    \end{tabular}
    }
    \label{tab:compressed_mbn2} \vspace{-4mm}
\end{table}

Finally, we demonstrate that NN-Mass can be used to compress even the most compact CNNs like MobileNets on ImageNet. Specifically, our model A (see Fig.~\ref{fig:atsc}(c,d)) is a significantly compressed version of mbn2-1.4 and can be directly identified using its NN-Mass (mbn2-1.4 and model A are very far in Fig.~\ref{fig:atsc}(c) but are clustered in  Fig.~\ref{fig:atsc}(d)). These results are summarized in Table~\ref{tab:compressed_mbn2}. As evident, our NN-Mass-based models allow up to $\bm{34\%}$ fewer MACs and $\bm{40\%}$ fewer \#Params than MobileNet-V2-1.4.

\vspace{-2mm}
\section{Conclusion}
\label{conc}
\vspace{-1mm}
We have proposed a new topological metric called \textit{NN-Mass} which quantifies how effectively information flows through DNNs. We have also established concrete theoretical relationships among NN-Mass, topological structure of DenseNet-type networks, and layerwise dynamical isometry that ensures faithful propagation of gradients through DNNs. 
Our training convergence MLP experiments have demonstrated that models with similar NN-Mass and width but different depths and number of parameters have similar training convergence and gradient flow properties like LDI. Our extensive experiments spanning DenseNets to MobileNets show that NN-Mass identifies models with similar accuracy, despite having a highly different number of parameters/FLOPS/layers. Finally, to show the practical applications of our work, we have exploited the closed-form equation of our NN-Mass metric to directly \textit{design} significantly compressed DenseNet-type CNNs (for CIFAR-10) and MobileNet-like CNNs (for ImageNet). 

Since topology is deeply intertwined with the gradient propagation, such topological metrics deserve major attention for future research. Another important venue for further work lies in the intersection of initialization and topology.

\vspace{-2mm}
\section*{Acknowledgement} 
We thank anonymous reviewers for helpful comments.  This  research  was  supported  in  part  by  U.S. National Science Foundation (NSF) under Grant CNS-2007284, Semiconductor  Research  Corporation  (SRC) under Grant SRC-2939.001, and Amazon AWS ML Research Program.

{\small
\bibliographystyle{ieee_fullname}
\bibliography{egbib}
}

\clearpage

\appendix

\section*{Supplementary Information: How does topology influence gradient propagation and model performance of deep networks with DenseNet-type skip connections?}

\section{DNNs/CNNs with random skip connections are Small-World Networks}\label{swaApp} 
\begin{figure*}[h]
\centering
\includegraphics[width=0.8\textwidth]{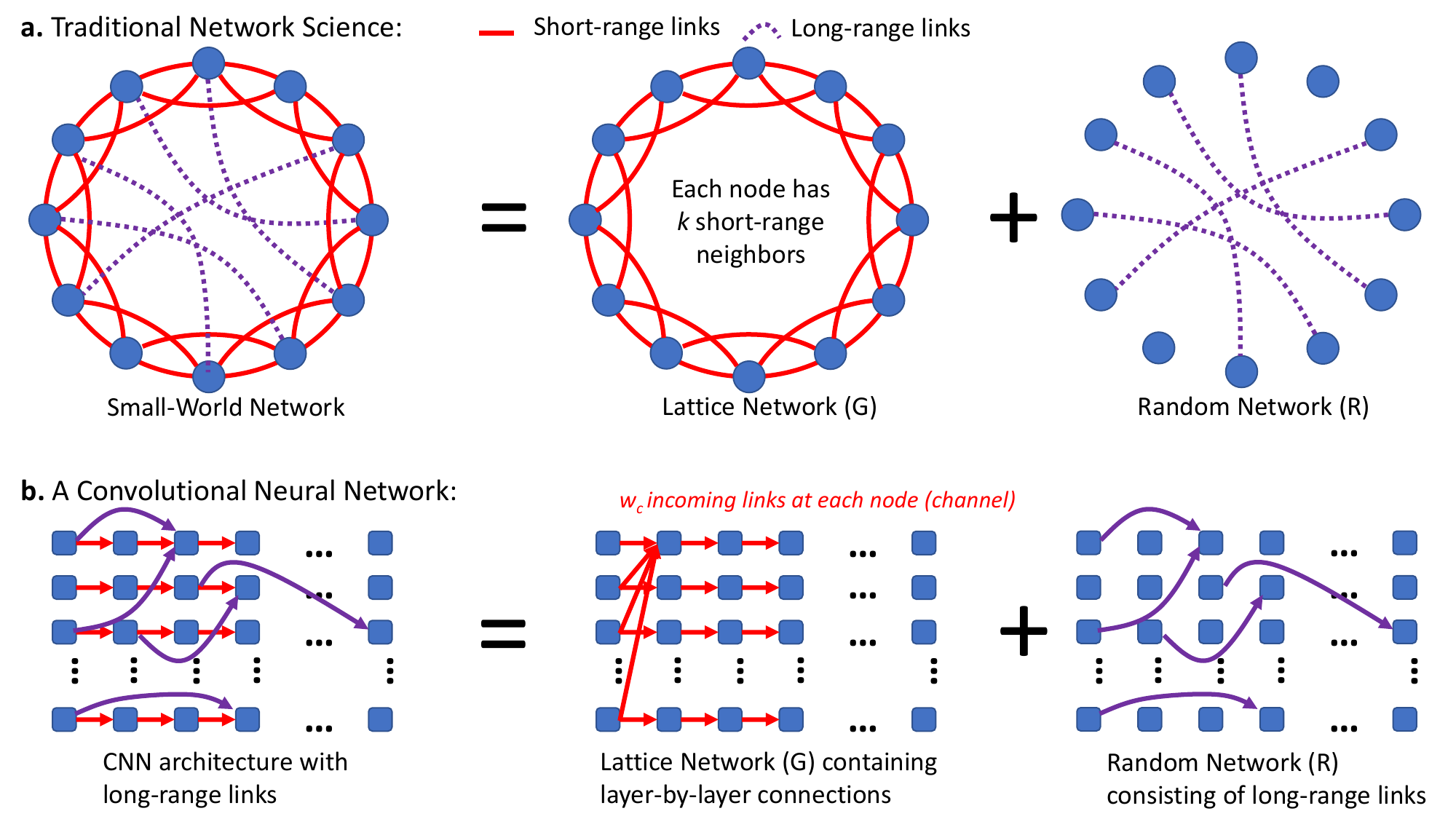}\vspace{-2mm}
	\caption{(a) Small-World Networks in traditional network science are modeled as a superposition of a lattice network ($\mathcal{G}$) and a random network $\mathcal{R}$~\cite{swnets, newman1999renormalization, monasson1999diffusion}. (b) A DNN/CNN with both short-range and long-range links can be similarly modeled as a random network superimposed on a lattice network. Not all links are shown for simplicity.}
\label{swa}
\end{figure*}

In network theory, a small world network is formed by superimposing a random network $\mathcal{R}$ on top of a lattice network $\mathcal{G}$ (see Fig.~\ref{swa})~\cite{monasson1999diffusion, newman1999renormalization}. As a result, these networks have both short-range and long-range links. 
Similarly, the DNNs/CNNs considered in our work have both short-range (due to layer-by-layer convolutions) and long-range links (due to random skip connections; see Fig.~\ref{fig2}(a)). This is illustrated in Fig.~\ref{swa}. 

\section{Derivation of Density of a Cell}\label{derNND}
Note that, the maximum number of neurons contributing skip connections at each layer in cell $c$ is given by $t_c$. Also, for a layer $i$, possible candidates for skip connections = all neurons up to layer ($i-2$) are $w_c(i-1)$ (see Fig.~\ref{fig2}(a)). Indeed, if $t_c$ is sufficiently large, initial few layers may not have $t_c$ neurons that can supply skip connections. For these layers, we use all available neurons for skip connections. Therefore, for a layer $i$, \#skip connections ($l_i$) is given by:
\begin{equation}
	l_i =
    \begin{cases}
      w_c(i-1) \times w_c & \text{if\ } t_c > w_c(i-1) \\
      t_c \times w_c        & \text{otherwise}
    \end{cases}
\label{lrlinks1}
\end{equation}
where, both cases have been multiplied by $w_c$ because once the neurons are randomly selected, they supply skip connections to all $w_c$ neurons at the current layer $i$ (see Fig.~\ref{fig2}(a)). Hence, for an entire cell, total number of neurons contributing skip connections ($l_c$) is as follows:
\begin{equation}
l_c = w_c \sum_{i=2}^{d_c-1}\text{min}\{w_c(i-1), t_c\}
\label{lc}
\end{equation}
On the other hand, the total number of possible skip connections within a cell ($L$) is simply the sum of possible candidates at each layer:
\begin{equation}
	\begin{aligned}
 		L &= \sum_{i=2}^{d_c-1} w_c(i-1)\times w_c= w_c^2 \sum_{i=2}^{d_c-1} (i-1) \\
 		&= w_c^2 [1 + 2 + \ldots + (d_c - 2)] \\
 		&= \frac{w_c^2 (d_c - 1) (d_c - 2)}{2}
	\end{aligned}
\label{L}
\end{equation}

Using Eq.~\ref{lc} and Eq.~\ref{L}, we can rewrite Eq.~\ref{rho1} as:
\begin{equation}
\rho_c = \frac{2\sum_{i=2}^{d_c-1}\text{min}\{w_c (i-1), t_c\}}{w_c (d_c - 1) (d_c - 2)}
\label{rho2}
\end{equation}
\qed

\section{Proof of Proposition 1}\label{prProp1App}
\setcounter{proposition}{0}
\begin{proposition}[\textbf{NN-Mass and average degree of the network (a topological property)}]
The average degree of a DenseNet-type deep network with NN-Mass $\nnMass$ is given by $\hat{k} = w_c+\nnMass/2$.
\end{proposition}

\noindent
\textit{Proof.} As shown in Fig.~\ref{swa}, deep networks with shortcut connections can be represented as small-world networks consisting of two parts: (\textit{i}) lattice network containing only the layer-by-layer links, and (\textit{ii}) random network superimposed on top of the lattice network to account for random skip connections. For sufficiently deep networks, the average degree for the lattice network will be just the width $w_c$ of the network. We consider the connections as undirected connections; hence each of the connection is counted only once.  The average degree of the randomly added skip connections $\rgKbar$ is given by:
\begin{equation}
	\begin{aligned}
 		\rgKbar &= \frac{\textit{Number of skip connections added by }\mathcal{R}}{\textit{Number of nodes}}\\
 		&= \frac{w_c\sum_{i=2}^{d_c -1}\text{min}\{w_c(i-1), t_c\}}{w_c d_c}\\
		&= \frac{\nnMass (d_c-1)(d_c-2)}{2d_c^2}\ \ \ \text{(Eq. \ref{mass1} for one cell, $N_c=1$)}\\
		&\approx \frac{\nnMass}{2}\ \ \ \ \ \  \text{(when } d_c >> 2\text{, \textit{e.g.}, for deep networks)}
	\end{aligned}
\label{krg1}
\end{equation}

\noindent
Therefore, average degree of the complete model is given by $w_c + \nnMass/2$. \qed

\section{Proof of Proposition 2}\label{prProp2App}
\begin{proposition}[\textbf{NN-Mass and LDI}]
Consider the case of deep linear networks with concatenation-type skip connections, where each layer is initialized using independently and identically distributed values with initialization variance $q$. For this setup, suppose we are given a small network $f_S$ (depth $d_S$) and a large network $f_L$ (depth $d_L$, $d_L>>d_S$), both with same initialization scheme, NN-Mass $\nnMass$, and width $w_c$. Then, the mean singular value of the initial layerwise Jacobian ($\mathbb{E}[\sigma]$) for both networks is bounded as follows: \vspace{-2mm}
\begin{equation*}
    \sqrt{q(w_c+m/2)}-\sqrt{qw_c} \leq \mathbb{E}[\sigma] \leq \sqrt{q(w_c+m/2)}+\sqrt{qw_c}\vspace{-2mm}
\label{prop2Bound}
\end{equation*}
That is, the LDI for both models does not depend on the depth if the initialization variance ($q$) for each layer is depth-independent (which is the case for many initialization schemes). Hence, for such networks, models with similar width and NN-Mass result in similar gradient properties, even if their depths and number of parameters are different.
\end{proposition}
\textit{Proof.} We now formally prove the above result under the assumption of deep linear networks~\cite{saxe2013exact,linear_network1,linear_network2,linear_network3}. As stated in the main text, layerwise Jacobians for linear networks follow a Gaussian distribution. We first prove the above result for a matrix $M\in \mathbb{R}^{H\times W}$ with $H$ rows and $W$ columns, where $H\geq W$, and all entries independently initialized with a Gaussian Distribution $\mathcal{N}(0, q)$. Towards the end of the proof, we will show how the results for matrix $M$ above apply to the correct layerwise Jacobians for a linear network ($\lj = \bm{W_i}$, where $\bm{W_i}$ is initialized as Gaussian with variance $q$).

To calculate the mean singular value of $M$, we perform Singular Value Decomposition (SVD) for matrix $M$:
\begin{equation*}
\begin{aligned}
    U\in \mathbb{R}^{H\times H},\mathrm{\Sigma}\in \mathbb{R}^{H\times W},V\in \mathbb{R}^{W\times W} =\text{SVD}(M)\\
    \mathrm{\mathrm{\Sigma}\in \mathbb{R}^{H\times W}}=Diag(\sigma_0,\sigma_1,...,\sigma_K)
\end{aligned}
\end{equation*}
Given the $i^{th}$ row vector $\vec{u_i}\in \mathbb{R}^H$ in $U$, and the $i^{th}$ row vector $\vec{v_i}\in \mathbb{R}^W$ in $V$, we use the following relations of SVD in our proof:
\begin{center}
$\sigma_i =\vec{u_i}^TM\vec{v_i}$ \\
$\vec{u_i}^T\vec{u_i} = 1$ \\
$\vec{v_i}^T\vec{v_i} = 1$  \\ 
\end{center}

It is hard to directly compute the mean singular value $\mathbb{E}[\sigma_i]$. To simplify the problem, consider $\sigma_i^2$ using the following product of SVD: 
\begin{equation}
\begin{aligned}
    M^TM&=U\Sigma V^TV\Sigma^TU^T=U(\Sigma\Sigma^T)U^T
\end{aligned}
\label{sigma_square}
\end{equation}
Consequently, the square of singular value ($\sigma_i^2$) of $M$ are the eigenvalues ($\lambda_i^{'}$) of $M^TM$:
\begin{equation}
    \sigma_i^2=\lambda_i^{'}
    \label{sigma_lambda}
\end{equation}

Mathematically, $\frac{1}{\sqrt{q}} M$ is a standard Gassian Random Matrix, i.e., all the elements of $\frac{1}{\sqrt{q}} M$ follow an i.i.d. standard Guassian Distribution $\mathcal{N} (0,1)$. From the theory of random matrix, the matrix $\frac{1}{q} M^TM$ is a Wishart ensemble \cite{wishart1928generalised}. 
Therefore, the distribution of the eigenvalues $(\lambda_1,\lambda_2,...,\lambda_i,...)$ of $\frac{1}{q} M^TM$ is Wishart Distribution. Then, for the Wishart Distribution, we know that the expectation of $(\lambda_1,\lambda_2,...,\lambda_i,...)$ \cite{wishart_textbook}:
\begin{equation}
    \mathbb{E}[(\lambda_1,\lambda_2,...,\lambda_i,...)] = (H,H,..,H,...)
    \label{expect_lambda}
\end{equation}

Clearly, $\lambda_i^{'} =q\lambda_i$. Combing Eq. \ref{sigma_lambda} and Eq. \ref{expect_lambda}, we get the following results:
\begin{equation}
    \mathbb{E}[\sigma_i^2]=qH
    \label{expect_h}
\end{equation}

\begin{figure}
    \centering
    \includegraphics[width=0.5\textwidth]{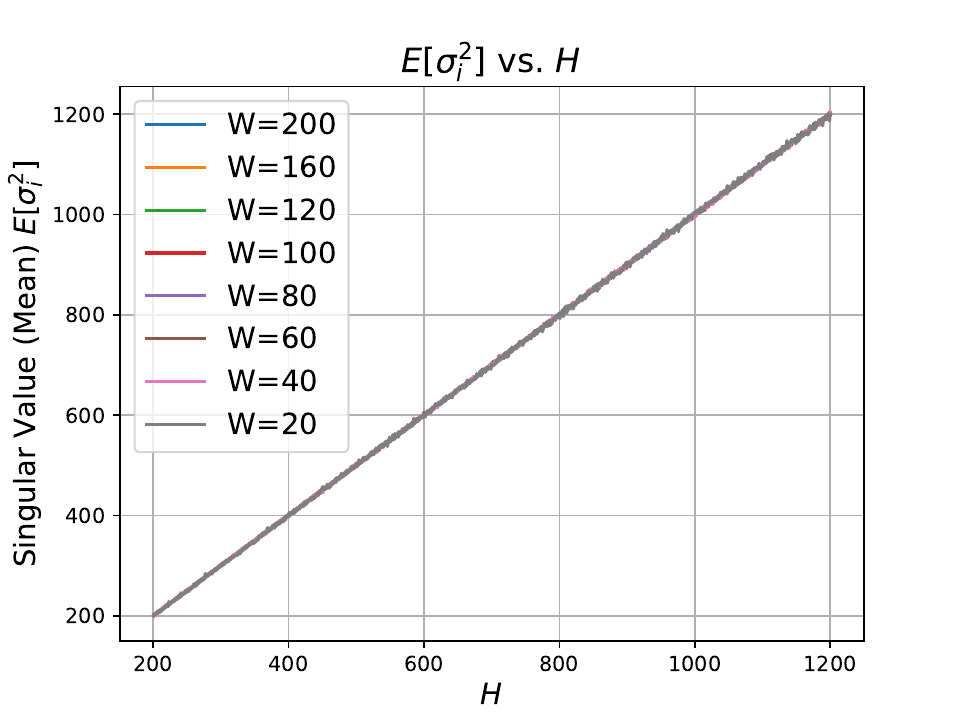} 
    \caption{Mean of square of singular value $\mathbb{E}[\sigma_i^2]$ only increases with $H$ while varying $W$ when $H\geq W$.}
    \label{sigma_h}
\end{figure}

Eq.~\ref{expect_h} states that, by keeping the same initialization variance $q$, for a Gaussian $M\in\mathbb{R}^{H\times W}$ with $H\geq W$, $\mathbb{E}[\sigma_i^2]$ is dependent on number of rows $H$, and does \textit{not} depend on W. To empirically verify this, we simulate several Gaussian matrices of widths $W\in\{20,40,80,120,160,200\}$ and $H\in[200,1200]$. We plot $\mathbb{E}[\sigma_i^2]$ \textit{vs.} $H$ in Fig. \ref{sigma_h}. As evident, the means of square of singular values $\mathbb{E}[\sigma_i^2]$ are nearly coinciding for different $W$, thereby showing that mean singular value indeed depends only on $H$.

While Eq.~\ref{expect_h} analytically shows the relationship between $\mathbb{E}[\sigma_i^2]$ and $H$, the LDI~\cite{sigprop} depends on the mean singular value $\mathbb{E}[\sigma_i]$ (and not its square). Although, the analytical expression for $\mathbb{E}[\sigma_i]$ is still an unsolved mathematical problem, it is still possible to provides its bounds. Using the theory of Marchenko–Pastur distribution, all singular values for matrix $M$ asymptotically lie in the interval $[\sqrt{qH}-\sqrt{qW}, \sqrt{qH}+\sqrt{qW}]$ \cite{marvcenko1967distribution,sigma_bound}. Therefore, the mean singular value also asymptotically lies in the interval $[\sqrt{qH}-\sqrt{qW}, \sqrt{qH}+\sqrt{qW}]$.

We now consider the initial layerwise Jacobian matrices ($\lj$) for the deep linear network scenario (i.e., $\lj = \bm{W_i}$, where $\bm{W_i}$ is initialized as Gaussian with variance $q$). As explained in the main paper, the layerwise Jacobians will theoretically have $(w_c+m/2, w_c)$ dimensions, where $w_c$ is the width of DNN and $m$ is the NN-Mass. That is, now $M = \lj$, $W = w_c$, and $H = w_c+m/2$. Hence, the above bounds for the mean singular value become:
\begin{equation}
    \sqrt{q(w_c+m/2)}-\sqrt{qw_c} \leq \mathbb{E}[\sigma] \leq \sqrt{q(w_c+m/2)}+\sqrt{qw_c}
    \label{prop2Bound}
\end{equation}

The above bound states that if the initialization variance $q$ is not dependent on depth (which is true for many initialization schemes), then, layerwise mean singular values for the given deep linear networks $f_S$ (depth $d_S$) and $f_L$ (depth $d_L, d_L>>d_S$) depends only on width $w_c$ and NN-Mass $m$. Therefore, for such deep networks, if $w_c$ and $m$ are the same, their layerwise dynamical isometry property (i.e., the mean singular values of initial layerwise Jacobians) has the same bounds. In other words, if two networks $f_S$ and $f_L$ have same width $w_c$ and NN-Mass $m$, they have similar gradient properties (i.e., LDI) even if they have significantly different depth and number of parameters. \qed

To empirically verify the Proposition~\ref{prop2} result, we plot the mean singular values as well as the bounds in~(\ref{prop2Bound}) for Gaussian distributed matrices of size ($w_c+m/2, w_c$) \textit{vs.} NN-Mass ($m$) in Fig.~\ref{fig2}(b) in the main paper. Clearly, the mean singular values for these simulated Jacobians fall within the above bounds. We will explicitly demonstrate in our experiments that NN-Mass is correlated with LDI for actual non-linear deep networks.

\section{CNN Details}\label{cnnSetupApp}

In contrast to our MLP setup which contains only a single cell of width $w_c$ and depth $d_c$, our CNN setup contains three cells, each containing a fixed number of layers, similar to prior works such as DenseNets~\cite{densenet}, Resnets~\cite{resnet}, \textit{etc}. However, topologically, a CNN is very similar to MLP. Since in a regular convolutional layer, channel-wise convolutions are added to get the final output channel (see Fig.~\ref{fig2}(c)), each input channel contributes to each output channel at all layers. This is true for both long-range and short-range links; this makes the topological structure of CNNs similar to our MLP setup shown in Fig.~\ref{fig2}(a) in the main paper (the only difference is that now each channel is a node in the network and not each neuron). 

In the case of CNNs, following the standard practice~\cite{vgg}, the width (\textit{i.e.}, the number of channels per layer) is increased by a factor of two at each cell as the feature map height and width are reduced by half. After the convolutions, the final feature map is average-pooled and passed through a fully-connected layer to generate logits. The width (\textit{i.e.}, the number of channels at each layer) of CNNs is controlled using a width multiplier, $wm$ (like in Wide Resnets~\cite{wrn} and Mobilenets~\cite{mobilenetV1}). Base \#channels in each cell is [16,32,64]. For $wm=2$, cells will have [32,64,128] channels per layer.

\section{Example: Computing NN-Mass for a CNN}\label{egApp}
\begin{figure*}[tb]
\centering
\includegraphics[width=\textwidth]{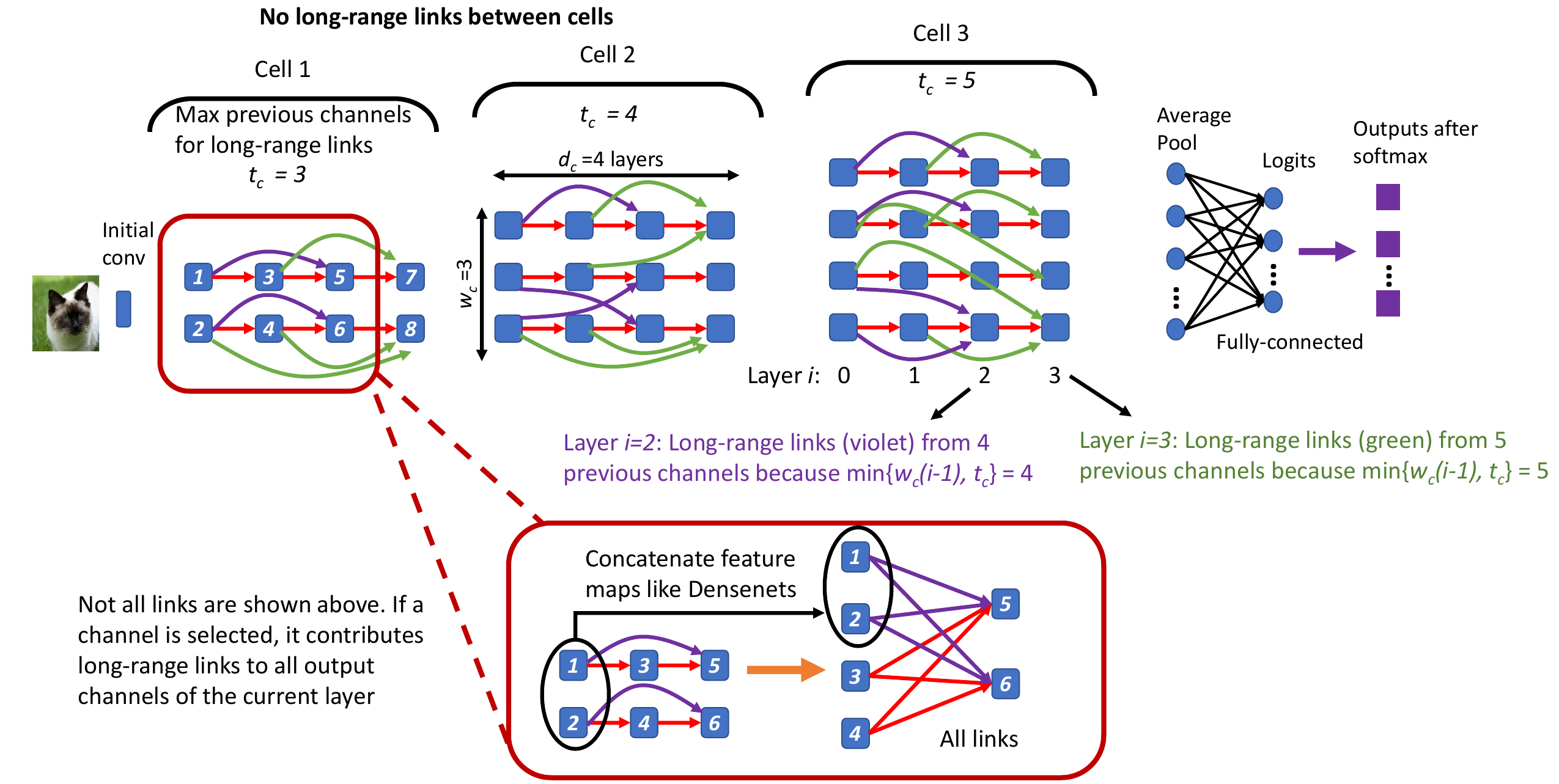}\vspace{-2mm}
	\caption{An example of CNN to calculate NN-Mass. Not all links are shown in the main figure for simplicity. The inset shows the contribution from all long-range and short-range links: The feature maps for randomly selected channels are concatenated at the current layer (similar to DenseNets~\cite{densenet}). At each layer in a given cell, the maximum number of channels that can contribute skip connections is given by $t_c$.}
\label{eg}
\end{figure*}

Given a CNN architecture shown in Fig.~\ref{eg}, we now calculate its NN-Mass. This CNN consists of three cells, each containing $d_c = 4$ convolutional layers. The three cells have a width, (\textit{i.e.}, the number of channels per layer) of 2, 3, and 4, respectively. We denote the network width as $w_c = [2,3,4]$. Finally, the maximum number of channels that can supply skip connections is given by $t_c = [3,4,5]$. That is, the first cell can have a maximum of three skip connection \textit{candidates} per layer (\textit{i.e.}, previous channels that can supply skip connections), the second cell can have a maximum of four skip connection candidates per layer, and so on. 
Moreover, as mentioned before, we randomly choose min$\{w_c(i-1), t_c\}$ channels for skip connections at each layer. The inset of Fig.~\ref{eg} shows how skip connections are created by concatenating the feature maps from previous layers.

Hence, using $d_c=4$, $w_c = [2,3,4]$, and $t_c = [3,4,5]$ for each cell $c$, we can directly use Eq.~\ref{mass1} to compute the NN-Mass value. Putting the values in the equations, we obtain $\nnMass = 28$. Consequently, the set $\{d_c, w_c, t_c\}$ can be used to specify the architecture of any CNN with concatenation-type skip connections. Therefore, to perform experiments, we vary $\{d_c, w_c, t_c\}$ to obtain architectures with different NN-Mass values.

\section{Complete Details of the Experimental Setup}\label{appExpDetails}
\subsection{MLP Setup}
We now explain more details on our MLP setup for the MNIST dataset. We create random architectures with different NN-Mass and \#Params by varying $t_c$ and $d_c$. Moreover, we just use a single cell for all MLP experiments. We fix $w_c=8$ and vary $d_c\in \{16,20,24,28,32\}$. For each depth $d_c$, we vary $t_c\in \{0,1,2,\ldots,14\}$. Specifically, for a given $\{d_c, w_c, t_c\}$ configuration, we create random skip connections at layer $i$ by uniformly sampling $\text{min}\{ w_c(i -  1),t_c\}$ neurons out of $w_c(i-1)$ activation outputs from previous $\{0, 1,\ldots,i-2\}$ layers. 

We train these random architectures on the MNIST dataset for 60 epochs with Exponential Linear Unit (ELU) as the activation function. Further, each $\{d_c,w_c, t_c\}$ configuration is trained five times with different random seeds. In other words, during each of the five runs of a specific $\{d_c, w_c, t_c\}$ configuration, the shortcuts are initialized randomly so these five models are not the same. This kind of setup is used to validate that NN-Mass is indeed a topological property of deep networks, and that the specific connections inside the random architectures do \textit{not} affect our conclusions. The results are then averaged over all runs: Mean is plotted in Fig.~\ref{mlp_mnist} and standard deviation, which is typically low, is also given in Fig.~\ref{mlp_mnist} caption. Overall, this setup results in many MLPs with different \#Params/FLOPS/layers.
\subsection{CNNs with DenseNet-type Skip Connections}
Much of the setup for creating concatenation-type skip connections in CNNs is the same as that for MLPs, except we have three cells instead of just one. As explained in Appendix~\ref{cnnSetupApp}, the width of the three cells is given as $wm\times[16,32,64]$, where $wm$ is the width multiplier. Note that, since we have three cells of different widths ($w_c$), $t_c$ also has a different value for each cell. The depth per cell $d_c$ is the same for all cells; hence, the total depth is given by $3d_c+4$. For instance, for 31-layer model, our $d_c=9$. For most of our experiments, we set the total depth of the CNN as $\{31,40,49,64\}$. Some of the experiments also use a total depth of $\{28,43,52,58\}$. 

Again, we conduct several experiments for different $\{d_c, w_c, t_c\}$ values which yield many random CNN architectures. The random skip connection creation process is the same as that in MLPs and, for CNN experiments, we have repeated all experiments three times with different random seeds. Specific numbers used for $\{d_c, w_c, t_c\}$ are given in Tables~\ref{expAll},~\ref{expAll1}, and~\ref{expAll3}. Each row in all tables represents a different $\{d_c, w_c, t_c\}$ configuration. Of note, all CNNs use ReLU activation function and Batch Norm layers.

For CNNs, we verify our findings on CIFAR-10 and CIFAR-100 image classification datasets. The learning rate for all models is initialized to 0.05 and follows a cosine-annealing schedule at each epoch. The minimum learning rate is 0.0 (see the end of Section~\ref{appDirRes} for details on how we fixed these hyper-parameter values). Similar to the setup in NAS prior works, the cutout is used for data augmentation. 

\subsection{MobileNet-v2 Setup}
We create random MobileNet-v2-like architectures with different NN-Mass, \#Params, and MACs by varying $N_c$ and width-multiplier. The standard MobileNet-v2 has $\{1,2,3,4,3,3,1\}$ inverted residual blocks with width (number of channels) $\{16,24,32,64,96,160,320\}$. Correspondingly, our searched compressed MobileNet-v2 has $\{2,3,4,5,4,3,3,1\}$ inverted residual blocks with the same width as standard MobileNet-v2. Furthermore, we sampled the width-multiplier as $\{0.15,0.35,0.6,0.75,0.9,1.0\}$.

As for the training process, we use the same training hyperparameters for all networks. We use SGD with $momentum=0.9$ and $weight-decay=4*10^{-5}$ as the optimizer. Moreover, we set the training epochs as 150, batch size as 256, and initial learning rate $lr_0=0.05$. For epoch $e$, the corresponding learning rate $lr_e=\frac{lr_0}{2}(1+cos(\frac{e-1}{150}\pi ))$. 
All models are trained in Pytorch on NVIDIA 1080-Ti, TitanXp, 2080-Ti, V100, and 3090 GPUs.  This completes the experimental setup.

\begin{table}[]
\caption{CNN architecture details (width multiplier = 2)\vspace{-3mm}}
\label{expAll}
\begin{center}
\scalebox{0.85}{
\begin{tabular}{llll}
\hline
\multicolumn{1}{|l|}{\begin{tabular}[c]{@{}l@{}}Number \\ of Cells\end{tabular}} & \multicolumn{1}{l|}{\begin{tabular}[c]{@{}l@{}}Max. Long-Range \\ Link Candidates ($t_c$)\end{tabular}}                                                                          & \multicolumn{1}{l|}{Depth} & \multicolumn{1}{l|}{Width Multiplier} \\ \hline \hline
\multicolumn{1}{|l|}{3}                                                          &
\multicolumn{1}{l|}{\begin{tabular}[c]{@{}l@{}}{[}10,35,50{]}\\ {[}20,45,75{]}\\ {[}30,50,100{]}\\ {[}40,60,120{]}\\ {[}50,70,145{]}\end{tabular}}           & \multicolumn{1}{l|}{31}    & \multicolumn{1}{l|}{2}     \\ \hline \hline
\multicolumn{1}{|l|}{3}                                                          & \multicolumn{1}{l|}{\begin{tabular}[c]{@{}l@{}}{[}20,40,70{]}\\ {[}30,50,100{]}\\ {[}40,80,125{]}\\ {[}50,105,150{]}\\ {[}60,130,170{]}\end{tabular}}        & \multicolumn{1}{l|}{40}    & \multicolumn{1}{l|}{2}     \\ \hline \hline
\multicolumn{1}{|l|}{3}                                                          & \multicolumn{1}{l|}{\begin{tabular}[c]{@{}l@{}}{[}25,50,90{]}\\ {[}35,80,125{]}\\ {[}50,105,150{]}\\ {[}70,130,170{]}\\ {[}90,150,210{]}\end{tabular}}       & \multicolumn{1}{l|}{49}    & \multicolumn{1}{l|}{2}     \\ \hline \hline
\multicolumn{1}{|l|}{3}                                                          & \multicolumn{1}{l|}{\begin{tabular}[c]{@{}l@{}}{[}30,80,117{]}\\ {[}50,110,150{]}\\ {[}70,140,200{]}\\ {[}90,175,250{]}\\ {[}110,215,300{]}\end{tabular}}    & \multicolumn{1}{l|}{64}    & \multicolumn{1}{l|}{2}     \\ \hline                   
\end{tabular}
}
\end{center}
\end{table}
\begin{table}[]
\caption{CNN architecture details (width multiplier = 1)\vspace{-3mm}}
\label{expAll1}
\begin{center}
\scalebox{0.85}{
\begin{tabular}{llll}
\hline 
\multicolumn{1}{|l|}{\begin{tabular}[c]{@{}l@{}}Number \\ of Cells\end{tabular}} & \multicolumn{1}{l|}{\begin{tabular}[c]{@{}l@{}}Max. Long-Range \\ Link Candidates ($t_c$)\end{tabular}}                                                                  & \multicolumn{1}{l|}{Depth} & \multicolumn{1}{l|}{Width Multiplier} \\ \hline \hline
\multicolumn{1}{|l|}{3}                                                          & \multicolumn{1}{l|}{\begin{tabular}[c]{@{}l@{}}{[}5,8,12{]}\\ {[}10,30,50{]}\\ {[}30,40,70{]}\\ {[}41,61,91{]}\\ {[}50,90,110{]}\end{tabular}}        & \multicolumn{1}{l|}{31}    & \multicolumn{1}{l|}{1}     \\ \hline \hline
\multicolumn{1}{|l|}{3}                                                          & \multicolumn{1}{l|}{\begin{tabular}[c]{@{}l@{}}{[}5,9,12{]}\\ {[}11,31,51{]}\\ {[}31,41,71{]}\\ {[}41,62,92{]}\\ {[}50,90,109{]}\end{tabular}}    & \multicolumn{1}{l|}{40}    & \multicolumn{1}{l|}{1}     \\ \hline \hline
\multicolumn{1}{|l|}{3}                                                          & \multicolumn{1}{l|}{\begin{tabular}[c]{@{}l@{}}{[}5,10,11{]}\\ {[}11,31,52{]}\\ {[}31,41,73{]}\\ {[}42,62,93{]}\\ {[}50,90,109{]}\end{tabular}}    & \multicolumn{1}{l|}{49}    & \multicolumn{1}{l|}{1}     \\ \hline \hline
\multicolumn{1}{|l|}{3}                                                          & \multicolumn{1}{l|}{\begin{tabular}[c]{@{}l@{}}{[}5,10,12{]}\\ {[}11,32,53{]}\\ {[}31,42,74{]}\\ {[}42,62,94{]}\\ {[}49,90,110{]}\end{tabular}} & \multicolumn{1}{l|}{64}    & \multicolumn{1}{l|}{1}     \\ \hline                   
\end{tabular}
}
\end{center}
\end{table}
\begin{table}[]
\caption{CNN architecture details (width multiplier = 3)\vspace{-3mm}}
\label{expAll3}
\begin{center}
\scalebox{0.85}{
\begin{tabular}{llll}
\hline
\multicolumn{1}{|l|}{\begin{tabular}[c]{@{}l@{}}Number \\ of Cells\end{tabular}} & \multicolumn{1}{l|}{\begin{tabular}[c]{@{}l@{}}Max. Long-Range \\ Link Candidates ($t_c$)\end{tabular}}                                                                  & \multicolumn{1}{l|}{Depth} & \multicolumn{1}{l|}{Width Multiplier} \\ \hline \hline
\multicolumn{1}{|l|}{3}                                                          & \multicolumn{1}{l|}{\begin{tabular}[c]{@{}l@{}}{[}10,30,50{]}\\ {[}40,60,90{]}\\ {[}70,90,130{]}\\ {[}100,120,170{]}\\ {[}130,150,210{]}\end{tabular}}        & \multicolumn{1}{l|}{31}    & \multicolumn{1}{l|}{3}     \\ \hline \hline
\multicolumn{1}{|l|}{3}                                                          & \multicolumn{1}{l|}{\begin{tabular}[c]{@{}l@{}}{[}11,31,51{]}\\ {[}42,62,92{]}\\ {[}72,93,133{]}\\ {[}103,123,173{]}\\ {[}133,153,212{]}\end{tabular}}     & \multicolumn{1}{l|}{40}    & \multicolumn{1}{l|}{3}     \\ \hline \hline
\multicolumn{1}{|l|}{3}                                                          & \multicolumn{1}{l|}{\begin{tabular}[c]{@{}l@{}}{[}11,31,52{]}\\ {[}43,63,93{]}\\ {[}73,95,135{]}\\ {[}104,124,176{]}\\ {[}134,154,214{]}\end{tabular}}    & \multicolumn{1}{l|}{49}    & \multicolumn{1}{l|}{3}     \\ \hline \hline
\multicolumn{1}{|l|}{3}                                                          & \multicolumn{1}{l|}{\begin{tabular}[c]{@{}l@{}}{[}12,32,52{]}\\ {[}44,64,95{]}\\ {[}76,96,136{]}\\ {[}106,126,178{]}\\ {[}135,156,216{]}\end{tabular}} & \multicolumn{1}{l|}{64}    & \multicolumn{1}{l|}{3}     \\ \hline                   
\end{tabular}
}
\end{center}
\end{table}

\clearpage
\section{Additional Results for DenseNet-type CNNs/MLPs}\label{appRes}
All results below are for DenseNet-type CNNs/MLPs.
\subsection{Results on synthetic data}\label{syntExApp}
\begin{figure}[h]
\centering
    \subfigure[Seg4]{
    \begin{minipage}[t]{0.5\linewidth}
    \centering
    \includegraphics[width=1\textwidth]{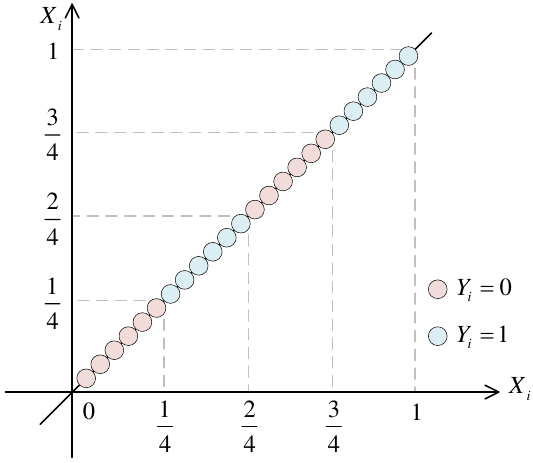}
    \end{minipage}%
    }%
    \subfigure[Circle4]{
    \begin{minipage}[t]{0.5\linewidth}
    \centering
    \includegraphics[width=1\textwidth]{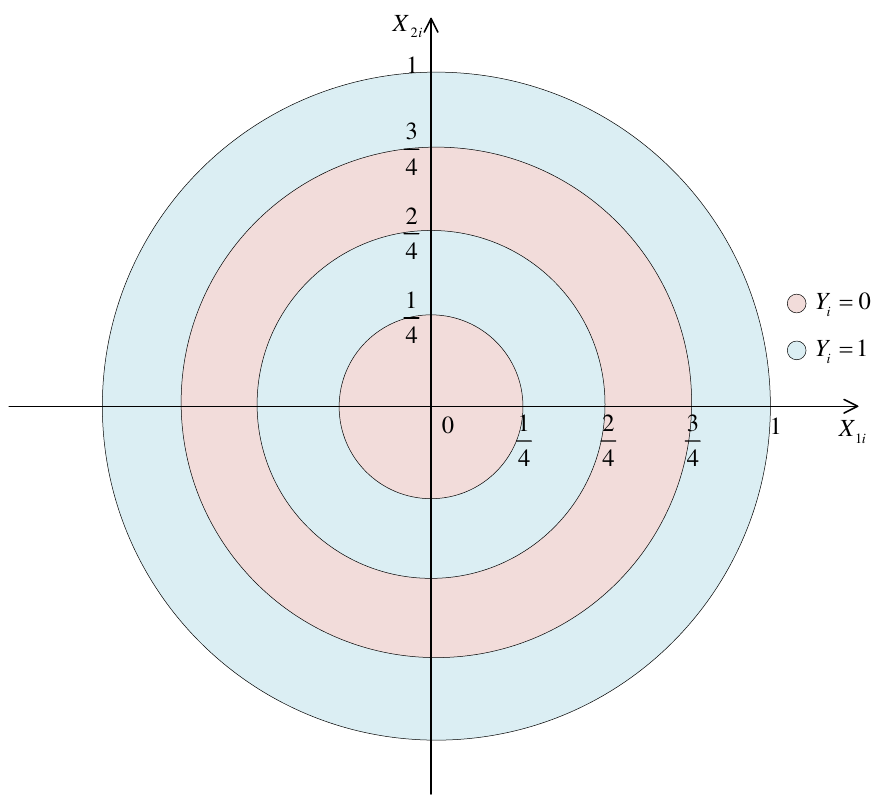}
    \end{minipage}%
    }%
\centering
\caption{Illustration of synthetic datasets Seg4 and Circle4: (a). Seg20 (Seg30) dataset is similar to Seg4, but divides the $[0,~1]$ range into 20 (30) segments. (b). Circle (or Circle20) dataset is similar to Circle4, but divides a unit circle into 20 concentric circles.}

\label{dataset_illu}
\end{figure}

In this section, we design a few synthetic experiments for MLP experiments to verify that our observations in Section~\ref{mlpRes} hold for diverse datasets. Specifically, we design three datasets -- Seg20, Seg30, and Circle20 (or just Circle). Fig.~\ref{dataset_illu}(a) illustrates the Seg4 dataset where the range $[0.~1]$ is broken into 4 segments. Similarly, Seg20 (Seg30) breaks down the linear line into 20 (30) segments. The classification problem has two classes (each alternate segment is a single class).

Fig.~\ref{dataset_illu}(b) shows the circle dataset where a unit circle is broken down into concentric circles (regions between circles make a class and we have two total classes). The details of these datasets are given in Table~\ref{table_synthetic}. Of note, we have used the ReLU activation function for these experiments (unlike ELU used for MNIST).
\begin{center}
    \begin{table*}[h!]
    \caption{Description of our generated Synthetic Datasets}
    \centering
        \begin{tabular}{|p{55pt}||p{270pt}|} 
        \hline
        Dataset name & Description: Training Set, $i\in [1,60000]$; Test Set, $i\in [1, 12000]$\\ 
        \hline\hline
        Seg20 & Feature: $[X_{i}, X_{i}]$, Label: $Y_{i}$,  $X_{i}= sample(\frac{1}{20}[\lfloor \frac{i}{20} \rfloor, \lfloor \frac{i}{20} \rfloor+1])$, $Y_{i}=\lfloor \frac{i}{20} \rfloor mod 2$\\ 
        \hline
        Seg30 & Feature: $[X_{i}, X_{i}]$, Label: $Y_{i}$,  $X_{i}= sample(\frac{1}{30}[\lfloor \frac{i}{30} \rfloor, \lfloor \frac{i}{30} \rfloor+1])$, $Y_{i}=\lfloor \frac{i}{30} \rfloor mod 2$\\ 
        \hline
        Circle (Circle20) & Feature: $[X_{1i}, X_{2i}]$, Label: $Y_{i}$,   $X_{1i}=L_{i}*cos(rand\_num), X_{2i}=L_{i}*sin(rand\_num), L_{i}=sample(\frac{1}{20}[\lfloor \frac{i}{20} \rfloor$, $Y_{i}=\lfloor \frac{i}{20} \rfloor mod 2$\\
        \hline
        \end{tabular}
    \label{table_synthetic}
    \end{table*}
\end{center}

For the above synthetic experiments, we once again conduct the following experiments: 
(\textit{i})~We explore the impact of varying \#Params and NN-Mass on the test accuracy. 
(\textit{ii})~We demonstrate how LDI depends on NN-Mass and \#Params. 

\begin{figure*}[tb]
\centering
    \subfigure[Linear: Seg=20]{
    \begin{minipage}[t]{0.333\linewidth}
    \centering
    \includegraphics[width=1\textwidth]{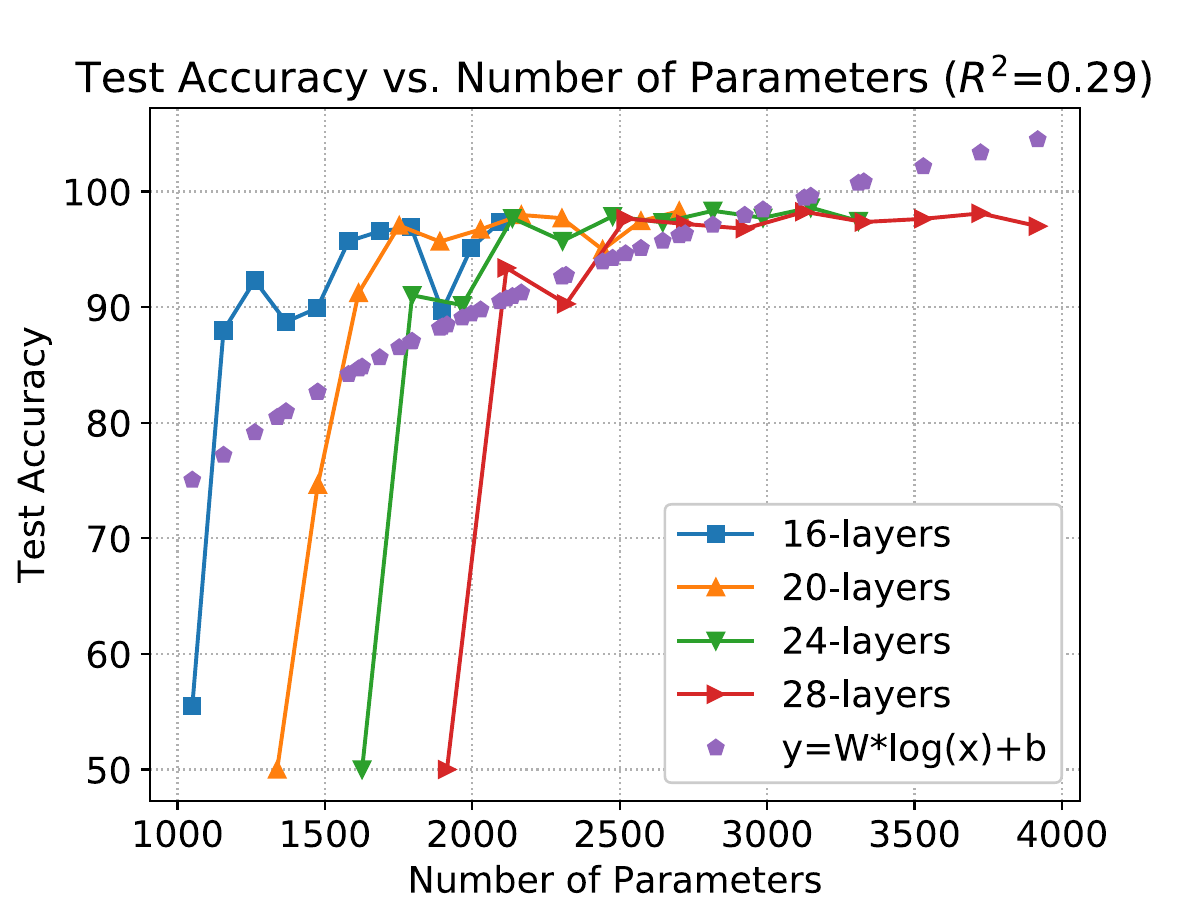}
    \end{minipage}%
    }%
    \subfigure[Linear: Seg=30]{
    \begin{minipage}[t]{0.333\linewidth}
    \centering
    \includegraphics[width=1\textwidth]{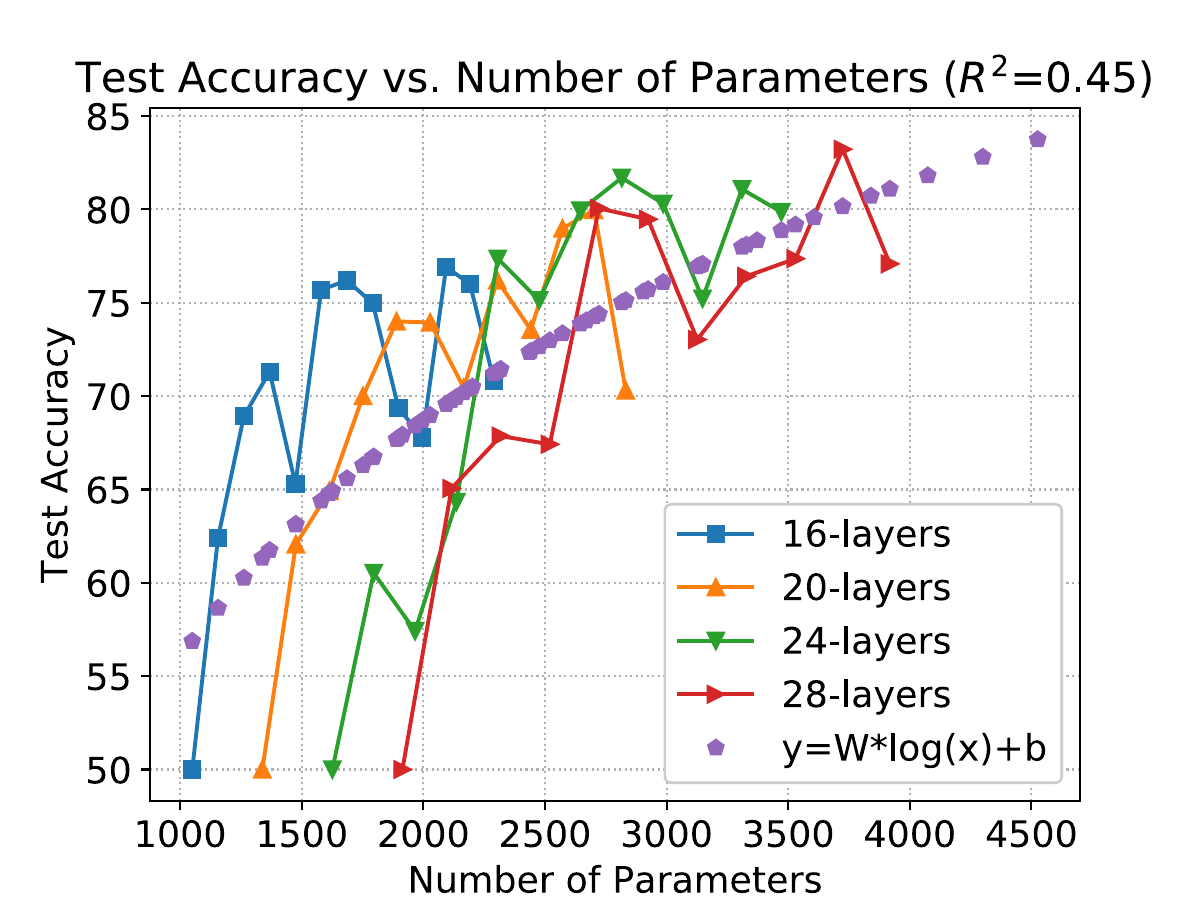}
    \end{minipage}%
    }%
    \subfigure[Circular: Circle20]{
    \begin{minipage}[t]{0.333\linewidth}
    \centering
    \includegraphics[width=1\textwidth]{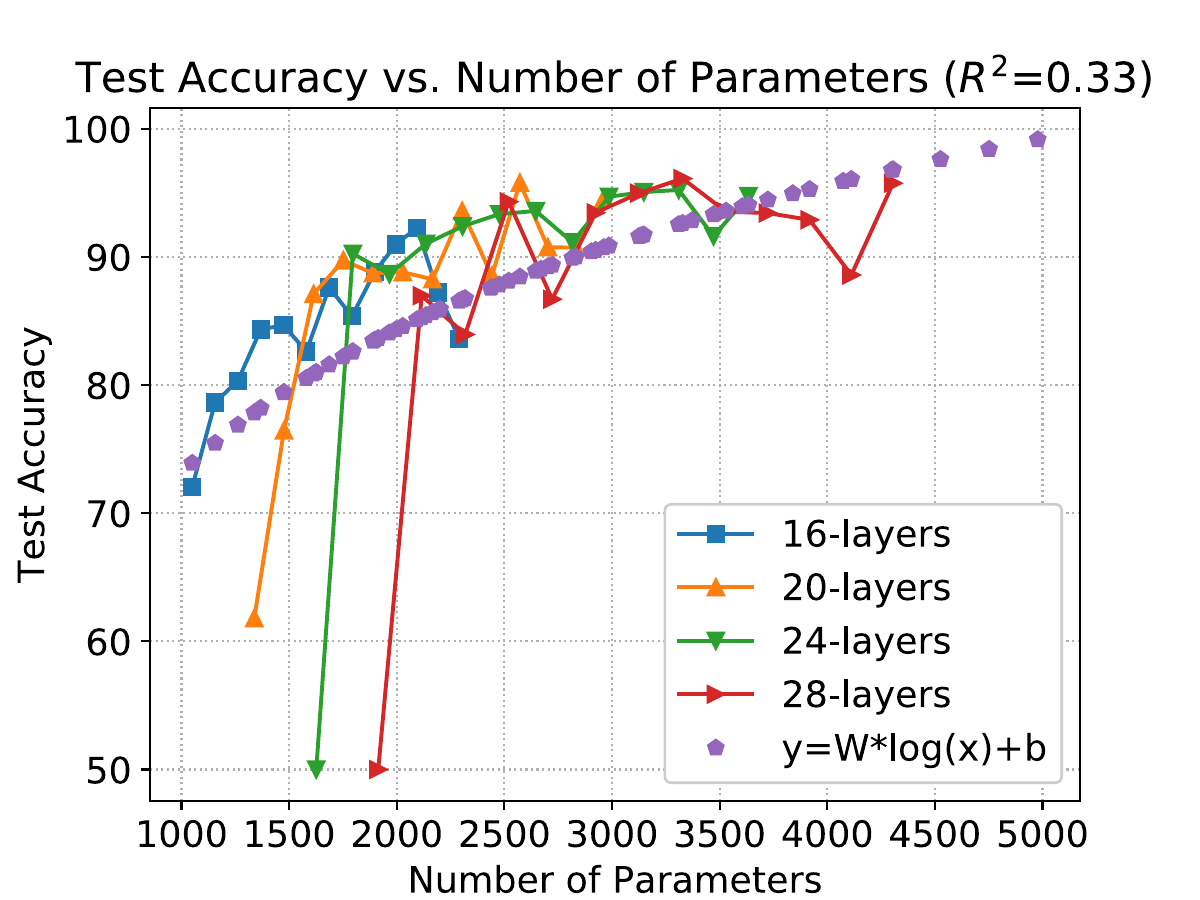}
    \end{minipage}%
    }%   
    
    \subfigure[Linear: Seg=20]{
    \begin{minipage}[t]{0.333\linewidth}
    \centering
    \includegraphics[width=1\textwidth]{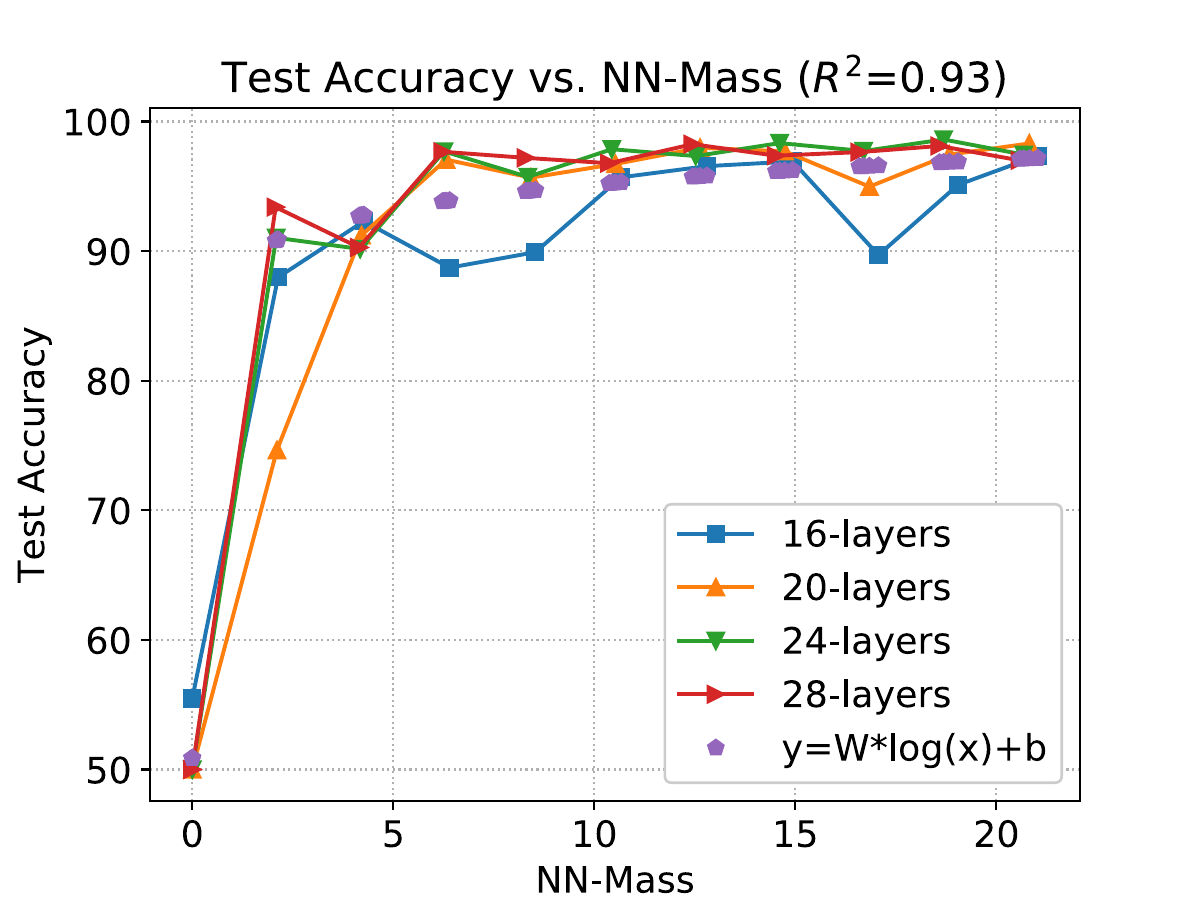}
    \end{minipage}%
    }%
    \subfigure[Linear: Seg=30]{
    \begin{minipage}[t]{0.333\linewidth}
    \centering
    \includegraphics[width=1\textwidth]{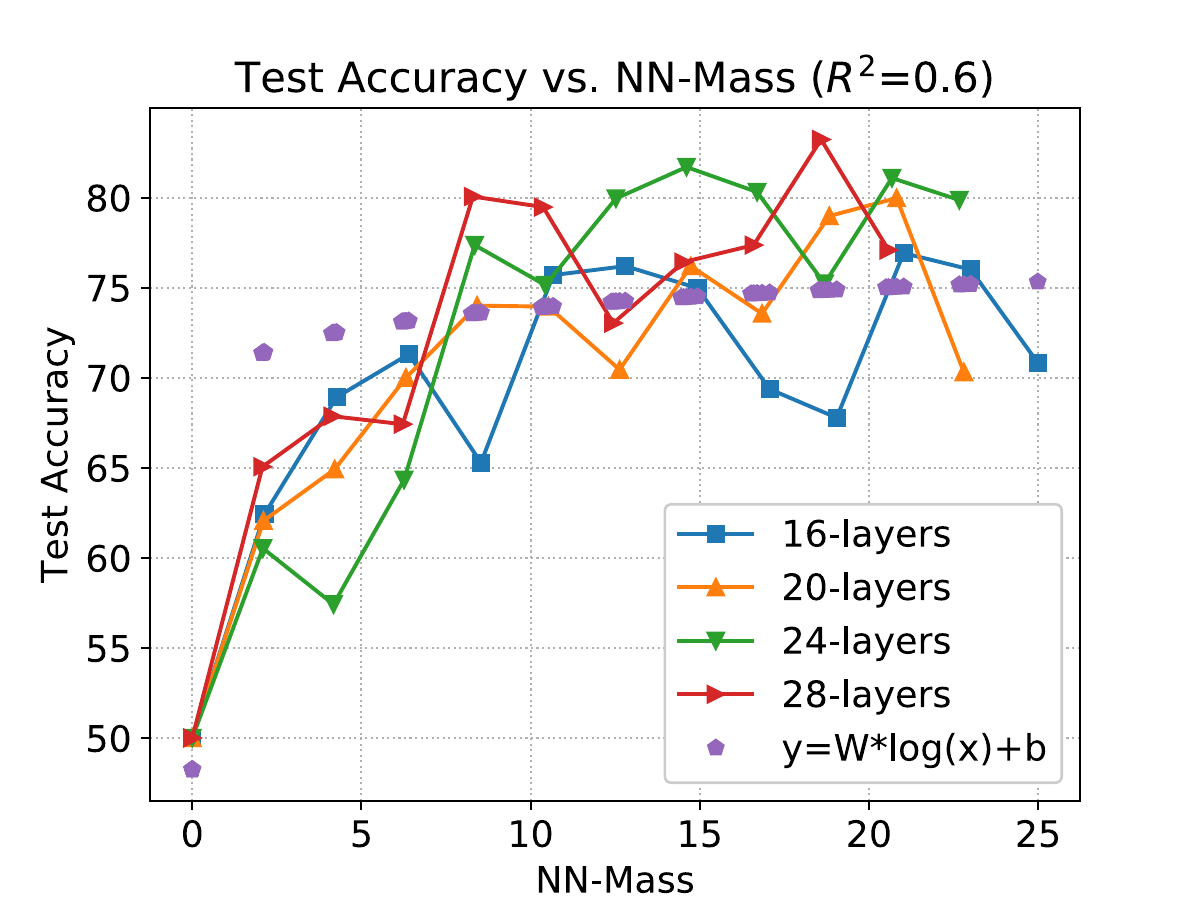}
    \end{minipage}%
    }%
    \subfigure[Circular: Circle20]{
    \begin{minipage}[t]{0.333\linewidth}
    \centering
    \includegraphics[width=1\textwidth]{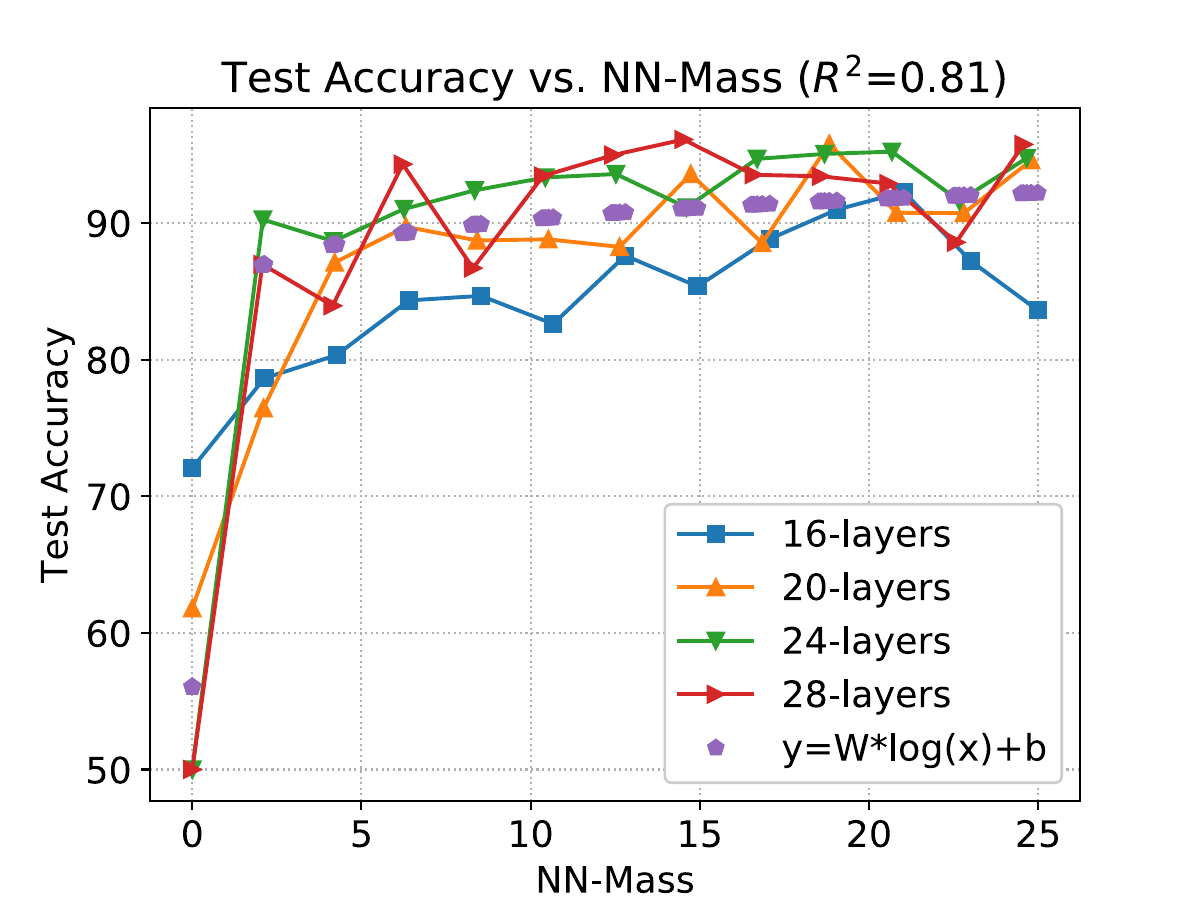}
    \end{minipage}%
    }%
\centering
\caption{Synthetic results: (a, b, c)~Models with different \#Params achieve similar test accuracy across all synthetic datasets. (d, e, f)~Test accuracy curves for the same set of models come closer together when plotted against NN-Mass.\vspace{-3mm}}

\label{mlp_synth}
\end{figure*}

\vspace{-1mm}
\paragraph{Test Accuracy}

As shown in Fig.~\ref{mlp_synth}(a, b, c) and Fig.~\ref{mlp_synth}(d, e, f), NN-Mass is a much better metric to characterize the model performance of DNNs than the number of parameters. Again, we quantitatively analyze the above results by generating a linear fit between test accuracy vs. log(\#Params) and log(NN-Mass). Similar to the MNIST case, our results show that $R^2$ of test accuracy vs. NN-Mass is much higher than that for \#Params.

\begin{figure}[tb]
\centering
    \subfigure{
    \begin{minipage}[t]{0.45\linewidth}
    \centering
    \includegraphics[width=1\textwidth]{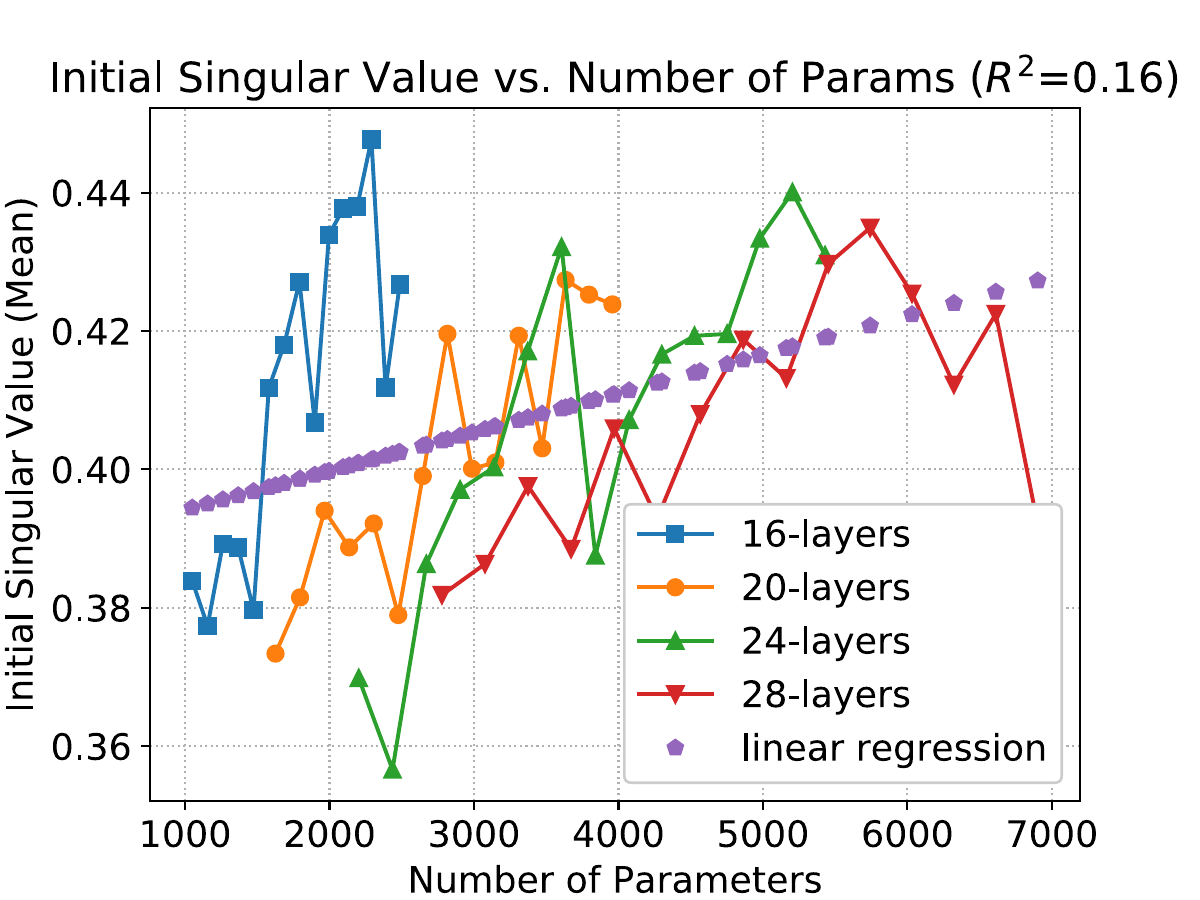}
    \end{minipage}
    }%
    \subfigure{
    \begin{minipage}[t]{0.45\linewidth}
    \centering
    \includegraphics[width=1\textwidth]{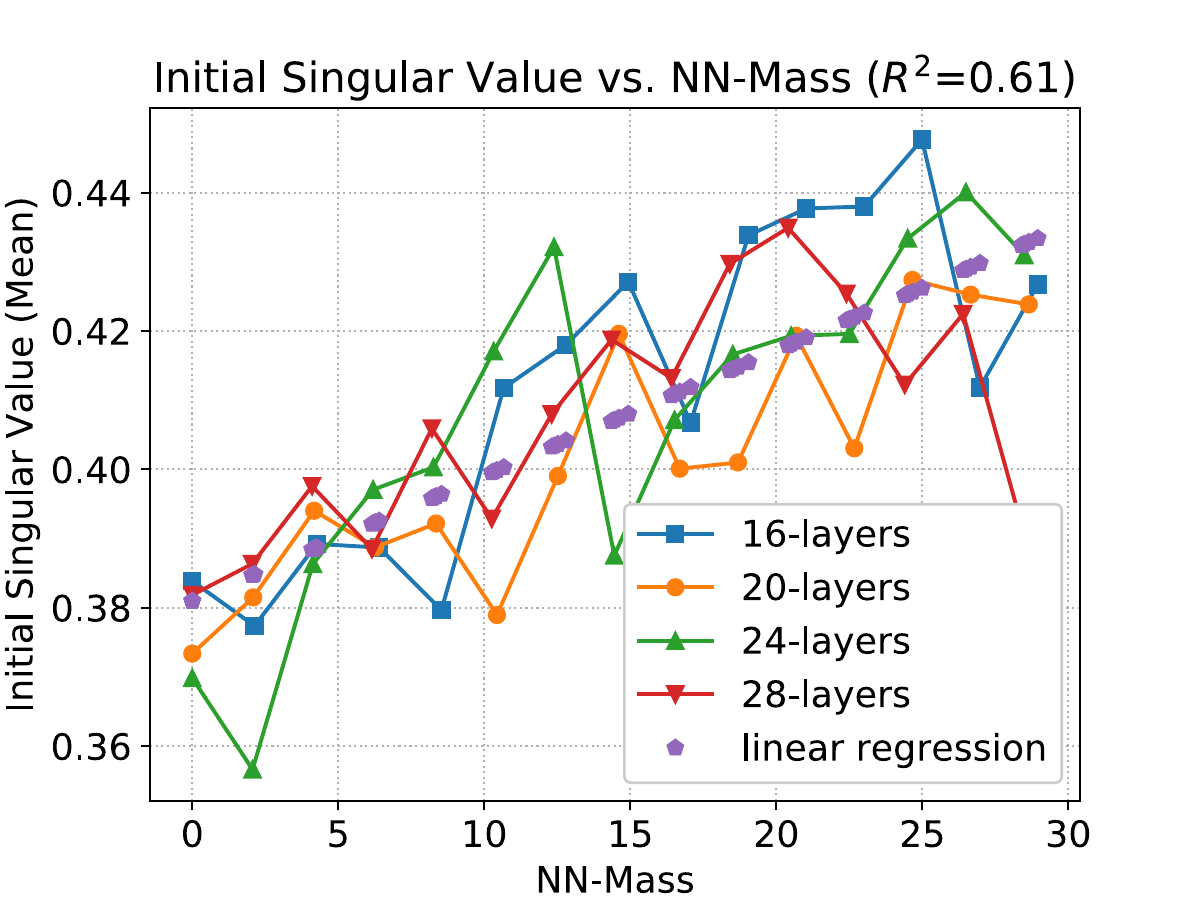}
    \end{minipage}
    }%
\centering
\caption{Synthetic results (Circle20 datasets):~Mean singular value of $\lj$ is much better correlated with NN-Mass than with \#Params.\vspace{-3mm}}
\label{isometry_synthetic}
\end{figure}

\paragraph{Layerwise Dynamical Isometry}
Fig. \ref{isometry_synthetic} shows the LDI results for the Circle20 dataset. Again, higher NN-Mass leads to higher initial singular value. Moreover, NN-Mass is better correlated with LDI than \#Params. Hence, this further emphasizes why networks with similar NN-Mass (instead of \#Params) result in a more similar model performance.

\subsection{Impact of Varying NN-Density}
NN-Density ($\rho_{avg}$) is defined as the average cell-density ($\rho_c$, see Definition~\ref{cellDensityDef}) across all cells in a DNN. 
As a baseline, we show that NN-Density cannot predict the accuracy of models with different depths. We train different deep networks with varying NN-Density (see Table~\ref{expAll} models in Appendix~\ref{appExpDetails}). Fig.~\ref{avd2} shows that shallower models with higher density can reach accuracy comparable to deeper models with lower density (which is quite reasonable since the shallower models are more densely connected compared to deeper networks, thereby promoting more effective information flow in shallower CNNs despite having significantly fewer parameters). However, NN-Density alone does not identify models (with different sizes/compute) that achieve similar accuracy: CNNs with different depths achieve comparable test accuracies at different NN-Density values (\textit{e.g.}, although a 31-layer model with $\rho_{avg}=0.3$ performs close to 64-layer model with $\rho_{avg}=0.1$, a 49-layer model with $\rho_{avg}=0.2$ already outperforms the test accuracy of the above 64-layer model; see models P, Q, R in Fig.~\ref{avd2}). Therefore, NN-Density alone is not sufficient.
\begin{figure}[tb]
    \centering
    \includegraphics[width=0.4\textwidth]{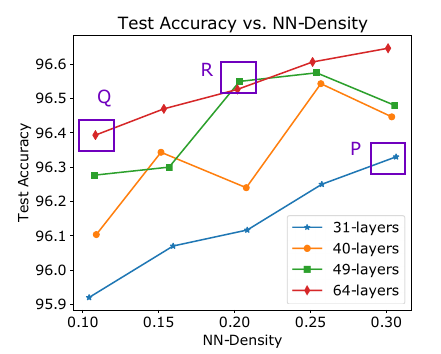}\vspace{-3mm}
    \caption{CIFAR-10 Width Multiplier $wm=2$: Shallower models with higher density can reach comparable accuracy to deeper models with lower density. This does not help since models with different depths achieve comparable accuracies at different densities.\vspace{-2mm}}
    \label{avd2}
\end{figure}

\subsection{Varying width multiplier on CIFAR-10}\label{diffWC10_App}
\begin{figure*}[!h]
    \centering
    \includegraphics[width=0.95\textwidth]{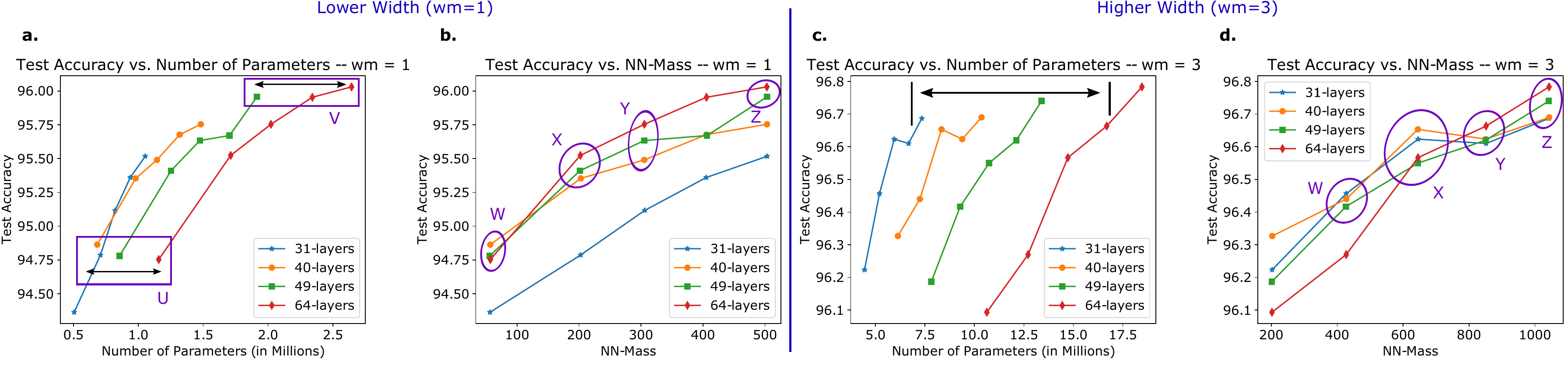}\vspace{-2mm}
    \caption{DenseNet-type CNNs for low- ($wm=1$) and high-width ($wm=3$) models: (a, b)~Many models with very different \#Params (boxes U and V) cluster into buckets W and Z (see also other buckets). (c, d)~For high-width, we observe a significantly tighter clustering compared to the low-width case. Results are reported as the mean of three runs (std. dev. $\sim$ $0.1\%$).}
    \label{avpavm13}
\end{figure*}

We now explore the impact of varying model width. In our DenseNet setup, we control the width of the models using \textit{width multipliers} ($wm$)\footnote{Base \#channels in each cell is [16,32,64]. For $wm=2$, cells will have [32,64,128] channels per layer.}~\cite{wrn, mobilenetV1}. The above results are for $wm=2$. For lower width CNNs ($wm=1$), Fig.~\ref{avpavm13}(a) shows that models in boxes U and V concentrate into the buckets W and Z, respectively (see also other buckets). Note that, the 31-layer models do not fall within the buckets (see blue line in Fig.~\ref{avpavm13}(b)). We hypothesize that this could be because the capacity of these models is too small to reach high accuracy. This does not happen for CNNs with higher width. Specifically, Fig.~\ref{avpavm13}(c) shows the results for $wm=3$. As evident, models with 6M-7M parameters achieve comparable test accuracy as models with up to 16M parameters (\textit{e.g.}, bucket Y in Fig.~\ref{avpavm13}(d) contains models ranging from $\{$31 layers, 6.7M parameters$\}$, all the way to $\{$64 layers, 16.7M parameters$\}$). Again, for all widths, the goodness-of-fit ($R^2$) for linear fit between test accuracy and log(NN-Mass) achieves high values (0.74-0.90 as shown in Fig.~\ref{varyW} in Appendix~\ref{appResR2C10}).

\subsection{$\bm{R^2}$ of CIFAR-10 Accuracy vs. NN-Mass}\label{appResR2C10}
\begin{figure*}[tb]
    \centering
    \includegraphics[width=0.92\textwidth]{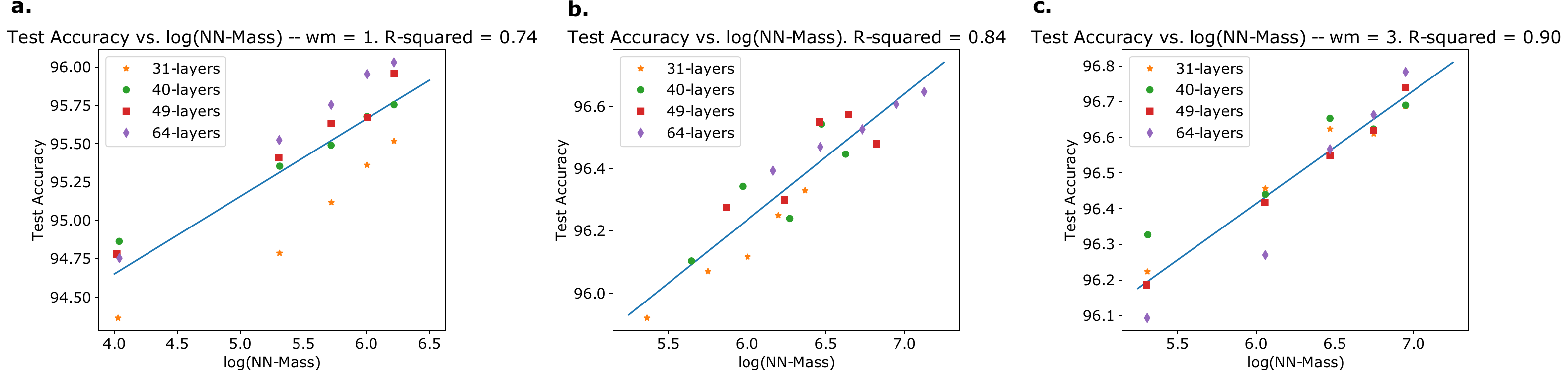}\vspace{-3mm}
    \caption{Impact of varying width of DenseNet-type CNNs: (a)~Width multiplier, $wm=1$, (b)~$wm=2$, and (c)~$wm=3$. As width increases, the capacity of small (shallower) models increases and, therefore, the accuracy-gap between models of different depths reduces. Hence, the $R^2$ for linear fit increases as width increases.\vspace{-3mm}}
    \label{varyW}
\end{figure*}

Fig.~\ref{varyW} shows the impact of increasing model widths on $R^2$ of linear fit between test accuracy and log(NN-Mass).

\subsection{Comparison between NN-Mass and Parameter Counting for CNNs}\label{appResComp}
\begin{figure*}[tb]
    \centering
    \includegraphics[width=0.9\textwidth]{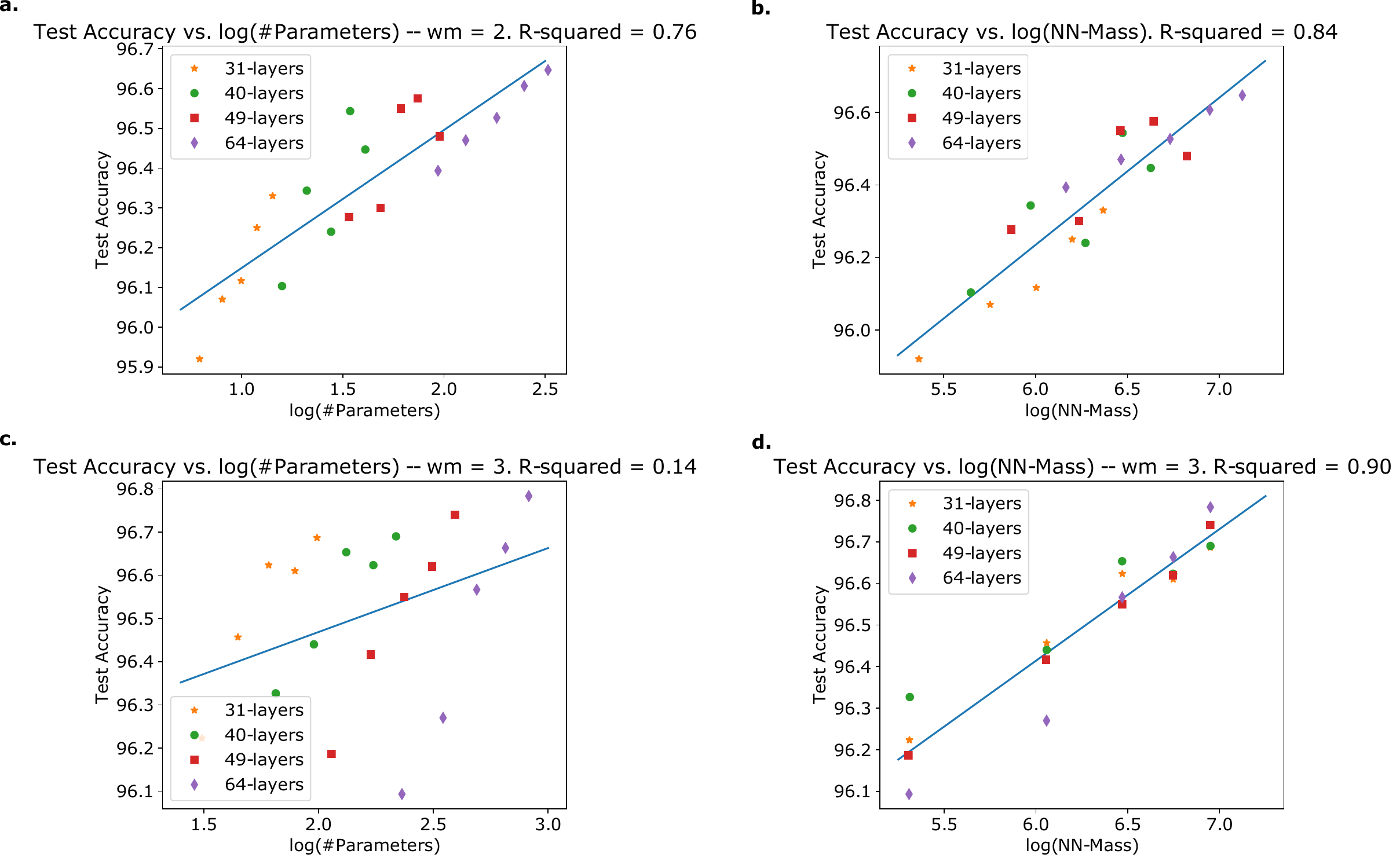}\vspace{-3mm}
    \caption{NN-Mass vs. parameter counting for DenseNet-type CNNs. (a)~For $wm=2$, log(\#Params) fits the test accuracy with an $R^2=0.76$. (b)~For the same $wm=2$ case, log(NN-Mass) fits the test accuracy with a higher $R^2=0.84$. (c)~For higher width ($wm=3$), parameter counting completely fails to fit the test accuracy of various models ($R^2=0.14$). (d)~In contrast, NN-Mass still fits the accuracies with a high $R^2=0.9$.}
    \label{genP}
\end{figure*}

For MLPs, we have shown that NN-Mass significantly outperforms \#Params for predicting model performance. For CNNs, we quantitatively demonstrate that while parameter counting can be a useful indicator of test accuracy for models with low width (but still not as good as NN-Mass), as the width increases, parameter counting completely fails to predict test accuracy. Specifically, in Fig.~\ref{genP}(a), we fit a linear model between test accuracy and log(\#Params) and found that the $R^2$ for this model is 0.76 which is slightly lower than that obtained for NN-Mass ($R^2=0.84$, see Fig.~\ref{genP}(b)). When the width multiplier of CNNs increases to three, parameter counting completely fails to fit the test accuracies of the models ($R^2=0.14$). In contrast, NN-Mass significantly outperforms parameter counting for $wm=3$ as it achieves an $R^2=0.90$. This demonstrates that NN-Mass is indeed a significantly stronger indicator of model performance than parameter counting.
\subsection{Results for CIFAR-100}\label{c100App}
\begin{figure*}[tb]
    \centering
    \includegraphics[width=0.93\textwidth]{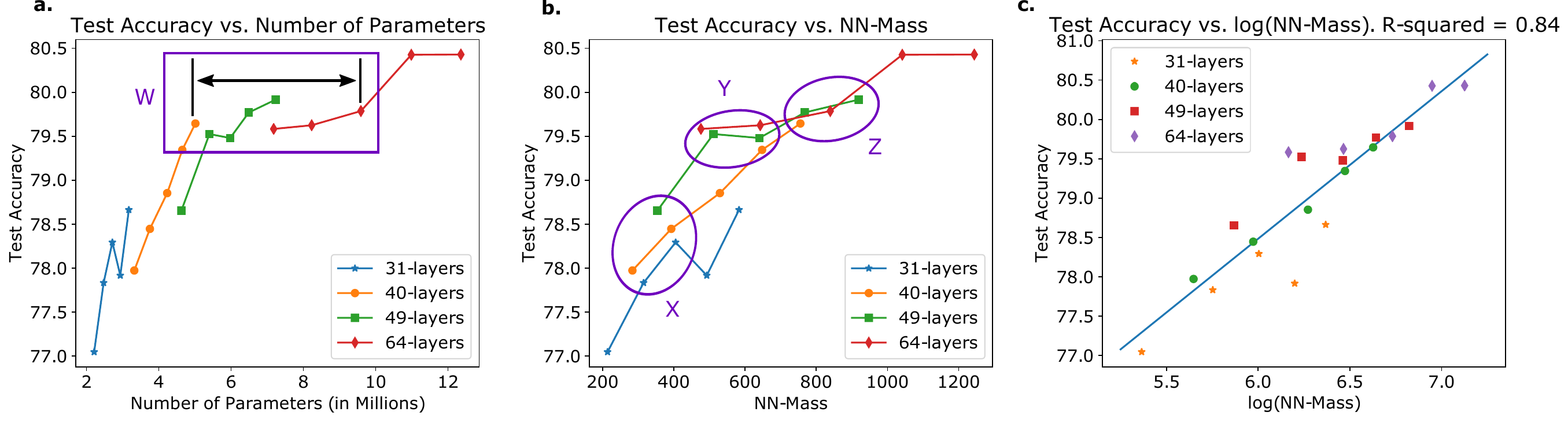}\vspace{-3mm}
    \caption{DenseNet-type CNNs: Similar results are obtained for CIFAR-100 ($wm=2$). (a) Models in box W have highly different \#Params but achieve similar accuracy. (b) These models get clustered into buckets Y and Z. (c) The $R^2$ value for fitting a linear regression model is $0.84$ which shows that NN-Mass is a good predictor of test accuracy. Results are reported as the mean of three runs (std. dev. $\sim$ $0.2\%$).}
    \label{avpavm2c100}
\end{figure*}

Results for CIFAR-100 dataset are shown in Fig.~\ref{avpavm2c100}. As evident, several models achieve similar accuracy despite having highly different number of parameters (\textit{e.g.}, see models within box W in Fig.~\ref{avpavm2c100}(a)). Again, these models get clustered together when plotted against NN-Mass. Specifically, models within box W in Fig.~\ref{avpavm2c100}(a) fall into buckets Y and Z in Fig.~\ref{avpavm2c100}(b). Hence, models that got clustered together for CIFAR-10 dataset, also get clustered for CIFAR-100. 
To quantify the above results, we fit a linear model between test accuracy and log(NN-Mass) and, again, obtain a high $R^2=0.84$ (see Fig.~\ref{avpavm2c100}(c)). Therefore, our observations hold true across multiple image classification datasets.
\begin{figure*}[htb]
\centering
\includegraphics[width=1\textwidth]{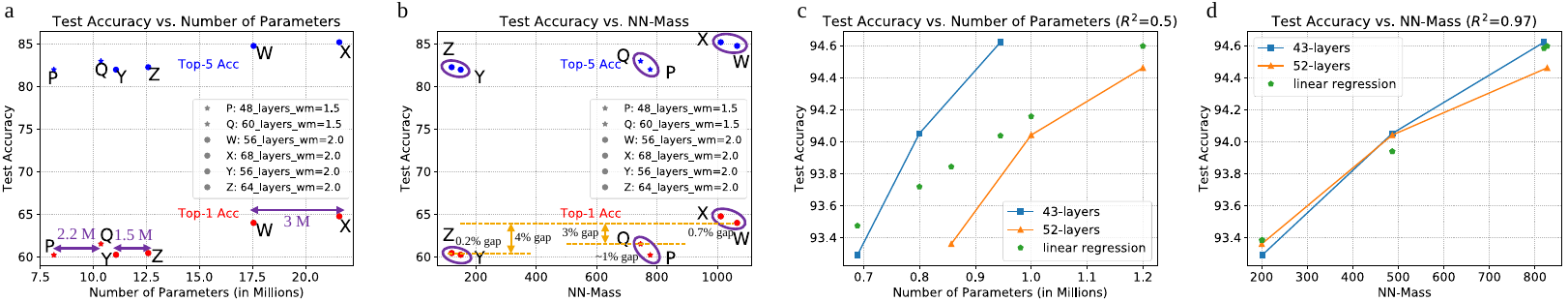}\vspace{-2.5mm}
    \caption{More results for DenseNet-type CNNs. (a,b) ImageNet: (a) Models \{P,Q\}, \{W,X\}, and \{Y,Z\} have very different \#Params but similar test accuracy. (b) When plotted against NN-Mass, the models with similar NN-Mass and accuracy cluster together. (c,d) CIFAR-10 with DSConv: Again, models with similar NN-Mass achieve similar accuracy but have quite different \#Params/layers.}\vspace{0mm}
\label{fig1}
\end{figure*}

\subsection{Results for ImageNet (DenseNet setup)}\label{dense_imagenet}
For ImageNet, we create several DenseNet-type CNNs containing four cells and total depth $\in\{48,56,60,64,68\}$ layers, and width multiplier $wm\in\{1.5,2\}$. Due to lack of resources, we trained these models on ImageNet dataset for 60 epochs. Fig. \ref{fig1}(a) shows the test accuracy of these CNNs vs. total \#Params, while Fig.~\ref{fig1}(b) shows the test accuracy vs. NN-Mass. As evident, although the model sizes are very different (e.g., model X is 3M parameters larger than model W; see other arrows also), the accuracy is quite similar. Once again, the models cluster together when plotted against NN-Mass (e.g., see clusters for models \{W,X\}, \{Y,Z\}, and \{P,Q\} in Fig.~\ref{fig1}(b)). Note that, the accuracies do \textit{not} saturate (similar to other CIFAR-10 and CIFAR-100 results in Fig.~\ref{avpavm2},~\ref{avpavm13},~\ref{avpavm2c100} and Fig.~\ref{fig1}(c,d) in next section): Cluster \{Y,Z\} achieves $4\%$ lower Top-1 accuracy (red points in Fig.~\ref{fig1}(a,b)) than cluster \{W,X\}, whereas within each cluster, the models are merely $0.2\%$ and $0.7\%$ away from each other. Same observation holds for Top-5 accuracy (blue points in Fig.~\ref{fig1}(a,b)). Finally, models \{P,Q\} cannot be compared in accuracy against \{Y,Z\} since they have different width (recall that Proposition 2 requires the models within the \textit{same} cluster to have both same width and NN-Mass). We have provided the $wm=1.5$ points to show that NN-Mass works for ImageNet across multiple widths. Hence, our ideas scale to the ImageNet dataset.

\subsection{Results for depthwise separable convolutions}\label{dense_ds_conv}
Recent works are heavily influenced by depthwise separable convolutions (DSConv). To demonstrate that NN-Mass works with DSConv, we take our current DenseNet setup (\textit{i.e.}, layer-by-layer convolutions with channels connected via random skip connections) and replace all convolutions with MobilenetV2 Expansion Blocks (1x1 conv $\rightarrow$ 3x3 DSConv $\rightarrow$ 1x1 conv)~\cite{mobilenetV2}. Random skip connections connect input channels across various Expansion Blocks. Fig. \ref{fig1}(c) shows test accuracy vs. \#Params of CNNs with DSConv on CIFAR-10. Again, even though many models have different \#Params, they achieve a similar  test accuracy. On the other hand, when the same set of models are plotted against NN-Mass, their test accuracy curves cluster together tightly, as shown in Fig. \ref{fig1}(d), with a significantly higher goodness-of-fit ($R^2=0.97$) than that for \#Params ($R^2=0.5$). This demonstrates that NN-Mass can be used to quantify topological properties of diverse/heterogeneous CNNs with regular convolutions, DSConv, pointwise conv. 

\subsection{Results for Floating Point Operations (FLOPS)}\label{flopsApp}
\begin{figure*}[tb]
    \centering
    \includegraphics[width=0.74\textwidth]{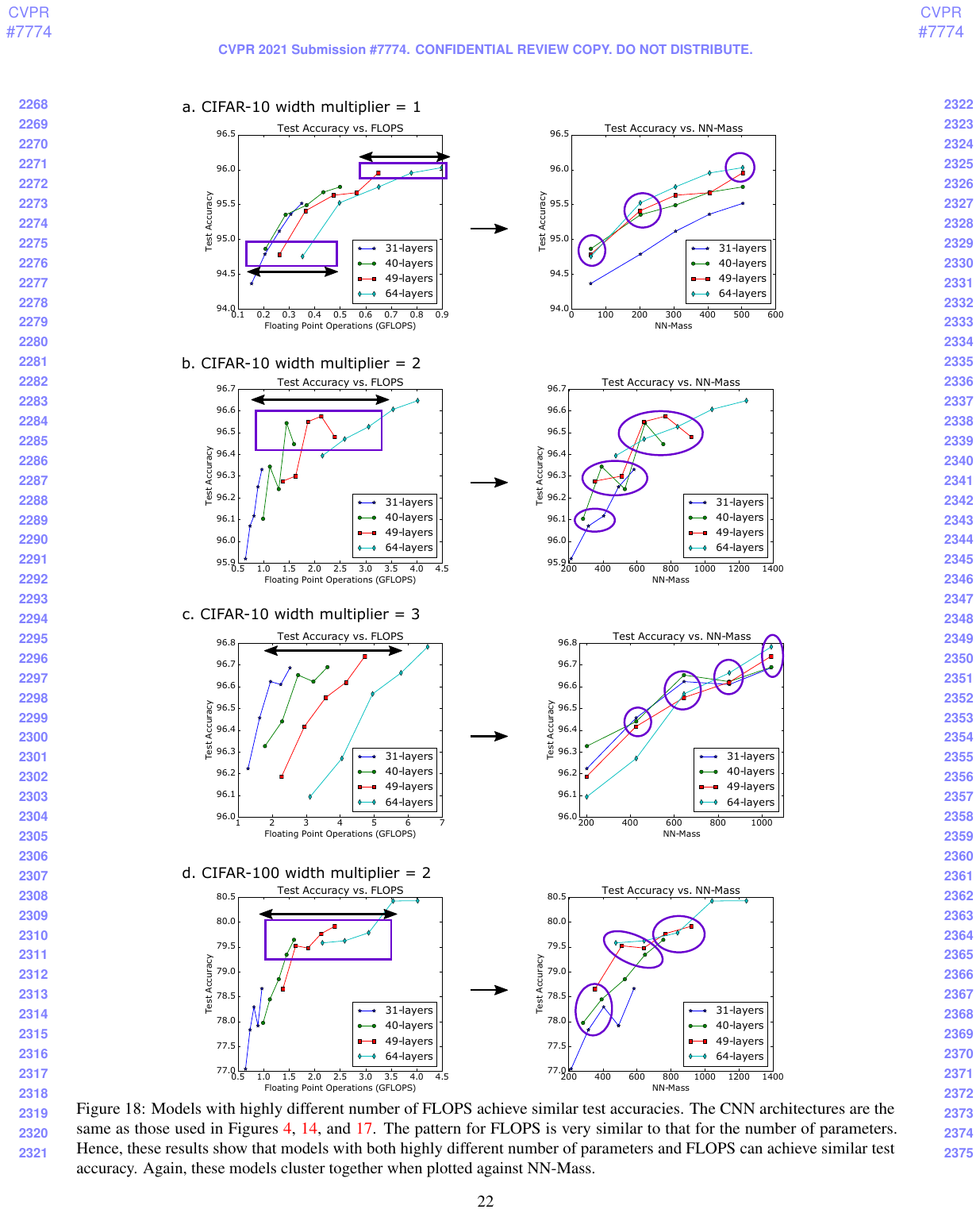}\vspace{-3mm}
    \caption{Models with highly different number of FLOPS achieve similar test accuracies. The CNN architectures are the same as those used in Figures~\ref{avpavm2},~\ref{avpavm13}, and~\ref{avpavm2c100}. The pattern for FLOPS is very similar to that for the number of parameters. Hence, these results show that models with both highly different number of parameters and FLOPS can achieve similar test accuracy. Again, these models cluster together when plotted against NN-Mass.}
    \label{flops}
\end{figure*}

All results for FLOPS (of CNN architectures in Tables~\ref{expAll},~\ref{expAll1}, and~\ref{expAll3}) are shown in Fig.~\ref{flops}. As evident, models with highly different number of FLOPS often achieve similar test accuracy. As shown earlier, many of these CNN architectures cluster together when plotted against NN-Mass.

\subsection{NN-Mass for directly designing compressed architectures}\label{appDirRes}
Our theoretical and empirical evidence shows that NN-Mass is a reliable indicator for models which achieve a similar accuracy despite having different number of layers and parameters. Therefore, this observation can be used for directly designing efficient CNNs as follows:
\begin{itemize}
\item 	First, train a reference big CNN (with a large number of parameters and layers) which achieves very high accuracy on the target dataset. Calculate its NN-Mass (denoted $\nnMass_L$).
\item 	Next, create a \textit{completely new and significantly smaller model} using far fewer parameters and layers, but with a NN-Mass ($\nnMass_S$) comparable to or higher than the large CNN. This process is very fast as the new model is created without any \textit{a priori} training. For instance, to design an efficient CNN of width $w_c$ and depth per cell $d_c$ and NN-Mass $\nnMass_S\approx \nnMass_L$, we only need to find how many skip connections to add in each cell. Since, NN-Mass has a closed form equation (\textit{i.e.}, Eq.~\ref{mass1}), a simple search over the number of skip connections can directly determine NN-Mass of various architectures. Then, we select the architecture with the NN-Mass close to that of the reference CNN. Unlike current manual or NAS-based methods, our approach does not require training of individual architectures during the search.
\item Since NN-Mass of the smaller model is similar to that of the reference CNN, our theoretical as well as empirical results suggest that the newly generated model will lose only a small amount of accuracy, while significantly reducing the model size. To validate this, we train the new, significantly smaller model and compare its test accuracy against that of the original large CNN. 
\end{itemize}

\paragraph{Directly designing compressed DenseNet-type CNNs for CIFAR-10.} We train our models for 600 epochs on the CIFAR-10 dataset (similar to the setup in DARTS~\cite{darts}). Table~\ref{massModC} (main paper) summarizes the number of parameters, FLOPS, and test accuracy of various CNNs. We first train two large CNN models of about 8M and 12M parameters with NN-Mass of 622 and 1126, respectively; both of these models achieve around $97\%$ accuracy. Next, we train three significantly smaller models: (\textit{i}) A 5M parameter model with 40 layers and a NN-Mass of 755, (\textit{ii}) A 4.6M parameter model with 37 layers and a NN-Mass of 813, and (\textit{iii}) A 31-layer, 3.82M parameter model with a NN-Mass of 856. 

We set the NN-Mass of our smaller models between 750-850 (\textit{i.e.}, within the 600-1100 range of the manually-designed CNNs). Interestingly, \textit{we do not need to train any intermediate architectures} to arrive at the above efficient CNNs. Indeed, classical NAS involves an initial ``search-phase'' over a space of operations to find the architectures~\cite{nasnet}. In contrast, our efficient models can be directly designed using the closed form Eq.~\ref{mass1} of NN-Mass (as explained in the beginning of this section), which does not involve any intermediate training or even an initial search-phase like prior NAS methods. As explained earlier, this is possible because NN-Mass can identify models with similar performance \textit{a priori} (\textit{i.e.}, without any training)! 

As evident from Table~\ref{massModC} (main paper), our 5M parameter model reaches a test accuracy of $97.00\%$, while the 4.6M (3.82M) parameter model obtains $96.93\%$ ($96.82\%$) accuracy on the CIFAR-10 test set. Clearly, all these accuracies are either comparable to, or slightly lower ($\sim 0.2\%$) than the large CNNs, while reducing \#Params/FLOPS by up to $3\times$ compared to the 11.89M-parameter/3.63G-FLOPS model. 
Moreover, DARTS~\cite{darts}, a competitive NAS baseline, achieves a comparable ($97\%$) accuracy with slightly lower 3.3M parameters. 
However, the search space of DARTS (like all other NAS techniques) is very specialized and utilizes many state-of-the-art innovations such as depth-wise separable convolutions~\cite{mobilenetV1}, dilated convolutions~\cite{dilated}, \textit{etc}. On the contrary, we use regular convolutions with only concatenation-type skip connections in our work and present a theoretically grounded approach. Indeed, our current objective is not to beat DARTS (or any other technique), but rather underscore the topological properties that should guide the efficient architecture design process. Ultimately, this theoretical knowledge (and its extensions to other kinds of networks) can help us drastically reduce the search space of NAS by directly removing architectures that are unlikely to improve accuracy.

\vspace{-2mm}
\paragraph{A note on hyper-parameter (\textit{e.g.}, initial learning rate) optimization.} Note that, throughout this work, we optimized the hyper-parameters such as initial learning rate for the largest models and then used the same initial learning rate for the smaller models. Hence, if these hyper-parameters were further optimized for the smaller models, the gap between the accuracy curves in Figures~\ref{avpavm13},~\ref{avpavm2c100},~\ref{flops}, \textit{etc.}, would reduce further (\textit{i.e.}, the clustering on NN-Mass plots would further improve). Similarly, the accuracy gap between compressed models and the large CNNs would reduce even more in Table~\ref{massModC} if the hyper-parameters were optimized for the smaller models as well. We did not optimize the initial learning rates, \textit{etc.}, for the smaller models as it would have resulted in an explosion in terms of number of experiments. Hence, since our focus is on topological properties of CNNs, we fixed the other hyper-parameters as described above. 
\end{document}